\documentclass[pdflatex,sn-basic,iicol]{sn-jnl}

\usepackage{graphicx}%
\usepackage{multirow}%
\usepackage{amsmath,amssymb,amsfonts}%
\usepackage{amsthm}%
\usepackage{mathrsfs}%
\usepackage[title]{appendix}%
\usepackage{xcolor}%
\usepackage{textcomp}%
\usepackage{manyfoot}%
\usepackage{booktabs}%
\usepackage{algorithm}%
\usepackage{algorithmicx}%
\usepackage{algpseudocode}%
\usepackage{listings}%
\usepackage{bm}%
\usepackage{array} 
\usepackage{graphicx} 
\usepackage{colortbl}
\definecolor{mygray}{gray}{.9}
\usepackage{cleveref} 
\crefname{equation}{Eq.}{Eqs.}
\Crefname{equation}{Eq.}{Eqs.}
\creflabelformat{equation}{#2(#1)#3}
\crefname{figure}{Fig.}{Figs.}
\Crefname{figure}{Fig.}{Figs.}
\crefname{section}{Sect.}{Sects.}    %
\Crefname{section}{Sect.}{Sects.}    %

\theoremstyle{thmstyleone}%

\theoremstyle{thmstyletwo}%

\theoremstyle{thmstylethree}%

\begin{document}

\title[Article Title]{CorrCLIPv2: Harnessing Unlabeled Images for Open-Vocabulary Semantic Segmentation}

\author[1]{\fnm{Dengke} \sur{Zhang}}\email{csdk@mail.scut.edu.cn}

\author*[1,2]{\fnm{Fagui} \sur{Liu}}\email{fgliu@scut.edu.cn}

\author*[2]{\fnm{Quan} \sur{Tang}}\email{tangq@pcl.ac.cn}

\author[1]{\fnm{C. L. Philip} \sur{Chen}}\email{Philip.Chen@ieee.org}

\affil[1]{\orgdiv{School of Computer Science and Engineering}, \orgname{South China University of Technology}, \orgaddress{\city{Guangzhou}, \postcode{510006}, \country{China}}}

\affil[2]{\orgdiv{Department of New Network}, \orgname{Pengcheng Laboratory}, \orgaddress{\city{Shenzhen}, \postcode{518000}, \country{China}}}

\abstract{Open-vocabulary semantic segmentation aims to classify pixels without being restricted to a fixed set of categories. Although Contrastive Language-Image Pre-training (CLIP) demonstrates remarkable potential for open-vocabulary recognition, its segmentation capability is impaired by noisy inter-class patch interactions. 
To this end, this work proposes a unified training-free object-centric framework, named CorrCLIPv2, from three complementary aspects of patch correlation, feature representation, and output regularization. Specifically, for patch correlation, we leverage class-agnostic mask generators to explicitly disentangle inter-class interactions. Moreover, we utilize spatially-coherent features from self-supervised models to compute attention scores to mitigate mask inaccuracies. Regarding feature representation, we devise a refinement mechanism to simultaneously strengthen spatial details and semantic representation. Within the output space, we unify predictions via voting in each mask to enhance spatial consistency.
To mitigate contextual bias and enhance prediction accuracy of objects, we further extend the framework by incorporating a cross-image object retrieval paradigm, which incurs minimal computational overhead. We build a memory bank of CorrCLIP-enhanced features from unlabeled images and implement a regime of multi-CLIP ensemble. By aggregating semantically consistent object features across diverse contexts retrieved from the memory bank, CorrCLIPv2 mitigates context-induced interference and obtains more robust predictions. 
Extensive experiments demonstrate that our approach establishes a new state-of-the-art on 10 standard benchmarks. Remarkably, it even surpasses fully supervised open-vocabulary methods across 22 multi-domain datasets.
Code is available at: \url{https://github.com/zdk258/CorrCLIP}.}


\maketitle

\begingroup
\renewcommand\thefootnote{}
\footnotetext{This preprint has not undergone peer review (when applicable) or any post-submission improvements or corrections. The Version of Record of this article is published in International Journal of Computer Vision, and is available online at \url{https://doi.org/10.1007/s11263-026-03027-3}.}
\endgroup

\section{Introduction}\label{sec1}

{O}pen-vocabulary semantic segmentation (OVSS) \citep{zhao2017open, openseg, ovsam} performs dense prediction by assigning semantic labels to image regions based on an open set of textual descriptions. In recent years, the Contrastive Language-Image Pre-training (CLIP) \citep{clip} model has emerged as a predominant paradigm for this task. Leveraging massive-scale supervision from image-text pairs, CLIP demonstrates unprecedented zero-shot generalization capabilities, thereby serving as a foundational backbone for contemporary OVSS frameworks.

\begin{figure}[t]
\centering
\includegraphics[width=1\linewidth]{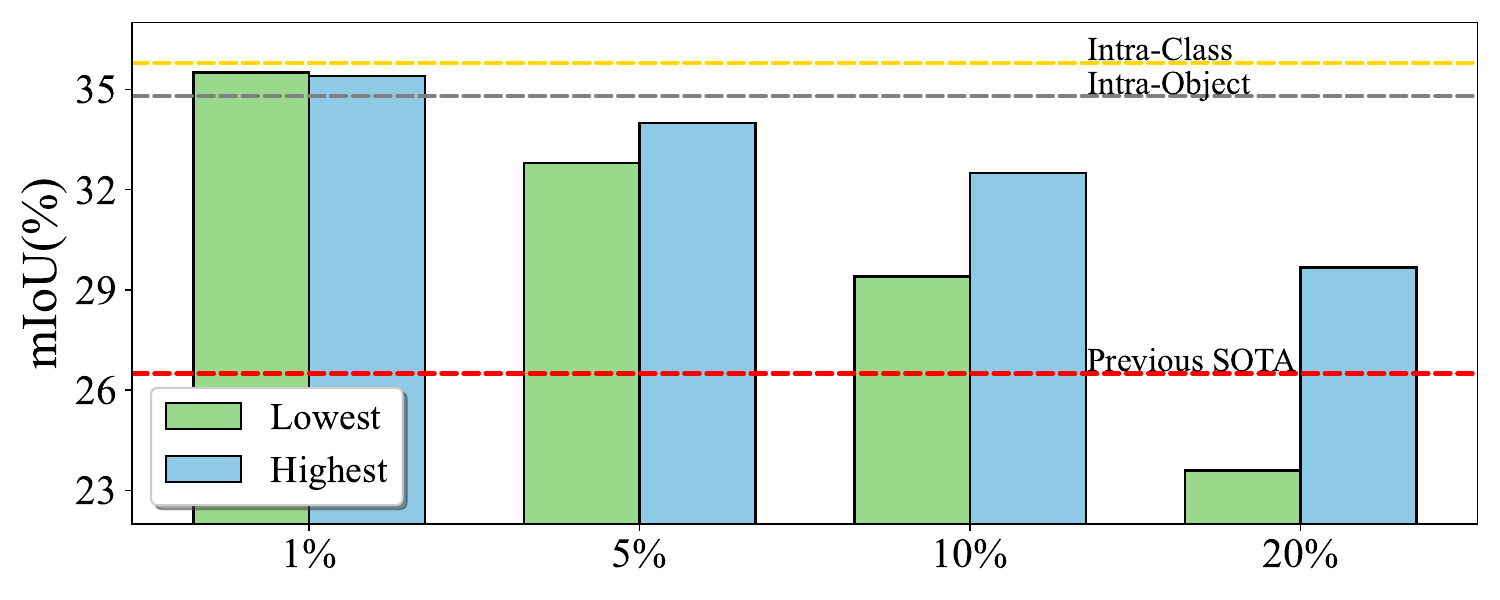}
\caption{Impact of intra-class and intra-object correlations and different proportions of inter-class correlations on CLIP's segmentation performance on COCO Stuff. ``Lowest'' and ``Highest'' denote the interactions with the least and most similar inter-class patches, respectively. Intra-Class and Intra-Object correlations achieve superior performance. In contrast, reintroducing inter-class interactions, even with high similarity, results in consistent degradation}
\label{fig_insight}
\end{figure}

Adapting CLIP to OVSS requires a fine-grained alignment between localized image patches and category embeddings. However, a fundamental discrepancy arises from CLIP’s contrastive pre-training, which prioritizes global information. As a result, distinct patches tend to share overly similar representations, undermining local discriminability. To enhance such local discriminability while preserving the pre-trained alignment with the textual feature space, training-free methods typically limit their intervention to the final Vision Transformer (ViT) \citep{vit} layer. Within this layer, the self-attention mechanism serves as the primary conduit for feature aggregation. Therefore, reshaping the attention patterns to restore local discriminability becomes the key challenge.

Existing methods aim to enhance segmentation primarily by adjusting attention maps heuristically. Specifically, SCLIP \citep{sclip} and ClearCLIP \citep{lan2024clearclip} substitute the conventional query-key interaction with query-query or key-key attention, enabling patches to focus more strongly on themselves to preserve local feature distinctiveness. Alternatively, SC-CLIP \citep{sc-clip} and ResCLIP \citep{resclip} observe that lower layer attentions exhibit superior coherence compared to the final layer, and thus leverage these shallower patterns to refine the noisy attention maps in the final block. Moreover, ProxyCLIP \citep{proxyclip} and Trident \citep{trident} bypass CLIP's internal inconsistencies by computing attention maps using features from other Visual Foundation Models (VFMs) that possess higher coherence. Critically, these methods do not explicitly identify which types of patch correlations degrade segmentation performance, thereby limiting their potential for further improvement.

\begin{figure}[t]
\centering
\includegraphics[width=0.9\linewidth]{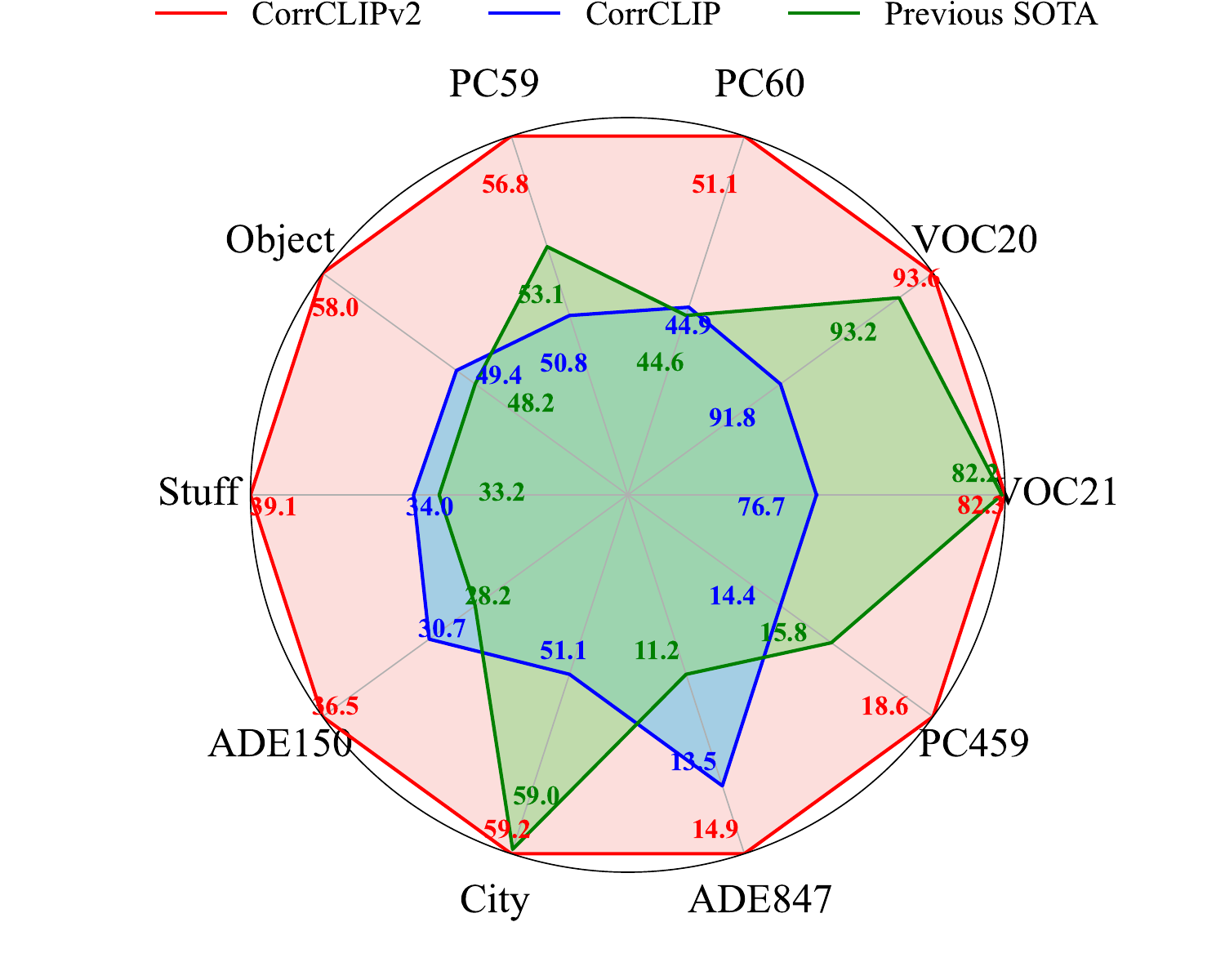}
\caption{Comparison of OVSS performance on ten standard benchmarks. CorrCLIP shows promising results. CorrCLIPv2 further improves it and sets a new state-of-the-art across all benchmarks}
\label{comparison_fig}
\end{figure}

To analyze this issue in depth, we conduct a breakdown analysis. For the segmentation task, ideal patch correlations in CLIP should maintain feature discriminability across classes and aggregate semantic information for semantic alignment. We categorize correlations into intra-class and inter-class types. Intuitively, intra-class correlations inherently satisfy both objectives. However, the impact of inter-class correlations presents a complex ambiguity. On one hand, aggregating information from distinct categories may introduce noise that compromises discriminability. On the other hand, it may provide useful contextual cues that facilitate alignment.
To resolve this ambiguity, we conduct a controlled investigation starting from a restricted setting of pure intra-class correlations. 
The empirical results are shown in \Cref{fig_insight}, suggesting a substantial improvement in CLIP's segmentation performance, which confirms the efficacy of intra-class aggregation. Additionally, intra-object correlations achieve results competitive with intra-class correlations. In contrast, we observe consistent degradation as inter-class correlations are reintroduced. Notably, even the most semantically similar inter-class patches are detrimental. These results suggest that, in the context of dense prediction, inter-class correlations primarily act as noise rather than providing useful contextual information.

Motivated by the aforementioned observation, we contend that explicitly regulating the interaction scope is imperative. To this end, we propose Scope Reconstruction that utilizes object masks generated by class-agnostic mask generators \citep{sam, sam2, entityseg} to strictly confine patch interactions within objects, thereby minimizing inter-class correlations. Furthermore, to avoid potential multi-class ambiguities in masks, we introduce Value Reconstruction to refine similarity measures via coherent features of Self-Supervised Learning (SSL) VFMs \citep{dinov2, dinov3, radiov2}, further suppressing the weight of irrelevant correlations. Beyond optimizing patch correlations, we present Feature Refinement, which introduces a spatial branch to recover spatial details from lower layers and a semantic branch utilizing mask class tokens to consolidate regional semantics. Finally, in the output space, we introduce Map Correction to unify predictions via voting in each mask, ensuring spatial consistency. These mechanisms effectively establish a robust object-centric framework, which reconstructs patch correlations, refines feature representations, and unifies prediction outputs within each object instance. 
While this framework improves object classification accuracy, inherent contextual bias may still cause object representations to deviate from their intrinsic semantics, leading to misclassification. This issue stems from the necessity of global contextual interaction in the preceding ViT layers of CLIP. Unlike the final layer where we can strictly confine the scope, early layers must aggregate information globally to form semantic concepts. Consequently, an object's embedding may be susceptible to interference from its surrounding context (e.g., a ``cat'' in a dark cage may be embedded closer to ``fence''). In such cases, relying solely on information from a single input image is insufficient to correct this semantic drift.

To this end, we extend our framework to a cross-image object retrieval paradigm, CorrCLIPv2, which incurs only marginal inference overhead. The core insight is that the feature representation of a single object instance may fluctuate due to contextual bias. In contrast, a semantic prototype distilled from multiple instances of the same class across diverse environments is more robust and reliable. Following this insight, we first construct an offline memory bank populated with CorrCLIP-enhanced features from a large collection of unlabeled data. A key challenge here is accurately linking the query object to semantically consistent support objects. To address this challenge, we introduce VFM-Indexed Semantic Retrieval (VISR). Unlike CLIP, SSL-based VFMs produce visually discriminative representations at the fine-grained object level, where intra-class features maintain high affinity, whereas inter-class features are well-separated. By utilizing VFM features as a stable anchor to retrieve a visual support set and aggregating their corresponding CorrCLIP features, VISR effectively marginalizes out the context-induced variance, stabilizing predictions under complex backgrounds.

Moreover, we exploit the decoupled design of our retrieval mechanism to enable a scalable approach for open-ended model integration, proposing a Decoupled Multi-CLIP Ensemble strategy. Unlike traditional ensembles that incur linear computational costs with each added model, our framework decouples the costly feature encoding of different CLIP variants, which is performed offline, from the online inference stage. Consequently, we can effectively fuse complementary knowledge dimensions from multiple pre-trained models with negligible marginal inference overhead, improving robustness without compromising efficiency.

Our contributions are summarized as follows:
\begin{itemize}
    \item We conduct a breakdown analysis revealing that inter-class correlations predominantly manifest as destructive noise rather than constructive context in CLIP-based segmentation. 

    \item Based on the above observation, we propose a unified training-free object-centric framework that reconstructs intrinsic patch correlations, refines feature representations, and unifies prediction outputs within each object instance. 

    \item To further enhance prediction accuracy, we extend our framework to a cross-image object retrieval paradigm, leveraging external support to mitigate contextual bias and incurring only marginal inference overhead.

    \item We index a large-scale memory bank using SSL-based VFMs to retrieve semantically consistent objects, and develop a decoupled multi-CLIP ensemble strategy that integrates diverse pre-trained models efficiently.

    \item Extensive experiments demonstrate the efficacy of CorrCLIPv2. It sets a new state-of-the-art on 10 standard benchmarks as shown in \Cref{comparison_fig} and outperforms fully supervised open-vocabulary counterparts on the multi-domain MESS benchmark comprising 22 datasets.

\end{itemize}

This work extends our preliminary study, CorrCLIP \citep{corrclip}, which has been accepted as an oral paper at ICCV'25. We evolve CorrCLIP into a cross-image paradigm with minimal inference overhead. 
The major improvements include:
(1) a retrieval-augmented framework that leverages unlabeled images to rectify contextual bias;
(2) the integration of SSL-based VFMs to construct a robust index, enabling the effective retrieval of semantically consistent support features;
(3) a decoupled ensemble strategy designed to efficiently leverage diverse pre-trained CLIP models;
(4) a substantial expansion of experiments on two challenging large-vocabulary benchmarks (ADE20K-847 \citep{ade}, Pascal Context-459 \citep{context}) and the MESS benchmark \citep{mess} with 22 multi-domain datasets, ensuring a more rigorous assessment.

\section{Related Work}\label{sec2}

\subsection{Open-Vocabulary Semantic Segmentation} 
Diverging from traditional semantic segmentation \citep{deeplab, mask2former, tang2025rethinking} methods that operate within a closed set of predefined labels, OVSS aims to segment images based on an open set of textual descriptions. The rapid evolution of this field is largely attributed to the emergence of Vision-Language Models (VLMs). Generally, existing OVSS methods can be categorized into training-based and training-free approaches.
Training-based approaches require additional fine-tuning using various forms of supervision, including mask annotations \citep{catseg,san,ovseg,dedos}, image-level labels \citep{clip-dinoier,seg4diff}, or image-text pairs \citep{Talk2DINO, dino.txt}. While these methods often yield superior performance on specific benchmarks, they tend to compromise the inherent generalization ability of VLMs as demonstrated in \citep{car}. Conversely, training-free methods circumvent the need for further training, thereby preserving the original zero-shot capabilities of VLMs.

Within the training-free landscape, a prominent research direction involves changing the attention mechanisms of CLIP. Several works \citep{maskclip,sclip,clip-srugery,lan2024clearclip, naclip, segearth} modify the final layer's attention, transitioning from a query-key interaction to a self-self mechanism to encourage patches to focus more on themselves. Others \citep{CLIPSeg, resclip, sc-clip, cliper} observe that lower layers often exhibit higher spatial coherence, thereby utilizing these layers to refine the noisy attention maps of the final layer. Despite these innovations, performance remains bottlenecked by CLIP’s incoherent features stemming from its fundamental training objective. To address this, recent studies \citep{proxyclip, cass, cliper, trident, freeda, lavg} have integrated auxiliary foundation models to bolster CLIP's segmentation capabilities. However, existing methods fail to explicitly regulate patch correlations. In this work, we introduce correlation reconstruction to explicitly reduce inter-class correlations, thereby significantly improving CLIP's segmentation performance.

\subsection{Vision-Language Pre-training} 
Vision-language pre-training (VLP) has emerged as a dominant paradigm for learning joint cross-modal representations. By harnessing weakly supervised learning on vast collections of image-text pairs, VLP enables models to capture intrinsic semantic associations between visual content and linguistic descriptions. While pioneering studies demonstrated the potential of this approach, their capabilities were inevitably limited by the scarcity of training data. A paradigm shift occurred with the advent of large-scale web-crawled datasets, enabling recent frameworks \citep{clip, align} to learn significantly more robust and transferable representations. Among these, CLIP \citep{clip} has established itself as the de facto standard. By leveraging contrastive learning to align visual and textual feature spaces, CLIP achieves remarkable zero-shot generalization. Building upon this foundation, a wave of subsequent research \citep{openclip, metaclip, pe, dfnclip} has further advanced the field by refining data curation strategies and optimizing training protocols. While CLIP lays the foundation for open-vocabulary semantic segmentation, its global image-text alignment objective is not inherently compatible with the pixel-level granularity required for segmentation, thereby requiring specific adaptations to bridge this gap.

\subsection{Vision Foundation Models} 
The primary objective of VFMs is to learn universal visual representations from large-scale data. Many VFMs build on self-supervised learning, which leverages data without human annotations. Early methods such as MAE \citep{mae} rely on reconstructive objectives that capture structural cues, while DINO \citep{dino} introduces self-distillation to obtain stronger semantic features.
This line of work is significantly advanced by DINOv2 \citep{dinov2}, whose large-scale curated training and improved optimization produce representations that capture intrinsic, high-level semantics. The recently developed DINOv3 \citep{dinov3} further strengthens this paradigm through enhanced data scaling and training stability, yielding consistently stronger generalization across recognition and dense prediction tasks.
Building on these advances, RADIO \citep{radiov2} distills knowledge from strong teacher models such as DINOv2 and CLIP, allowing it to inherit rich semantic representations and local details.
Benefiting from the global semantic coherence and local fine-grained details of SSL-based VFMs, we employ their features to suppress irrelevant correlations and construct indices to retrieve semantically consistent support features to stabilize predictions.

\subsection{Class-Agnostic Mask Generator} 
Semantic segmentation jointly learns category recognition and pixel-level delineation, requiring models to decide both what and where. This reliance on fixed label spaces limits generalization to open-world scenarios. Class-agnostic mask generators, by contrast, focus solely on mask prediction without semantic classification, making them easier to train and inherently more robust to unseen categories.
Large-scale vision models have recently advanced this paradigm. The Segment Anything Model (SAM) \citep{sam, sam2} learns from a massive image–mask corpus and employs a promptable architecture to produce high-quality, zero-shot masks, establishing a general foundation for open-world segmentation. EntitySeg \citep{entityseg, entityseg_pami} similarly targets the segmentation of generic entities rather than predefined classes, using strong visual priors to achieve class-agnostic object extraction.
Based on these advanced open-world mask generators, we construct a training-free object-centric OVSS framework and further extend it to a cross-image object retrieval paradigm.

\subsection{Memory Bank for OVSS} 
Memory bank has recently emerged as a promising paradigm to enhance OVSS. However, existing methods predominantly construct memory banks heavily anchored to textual guidance. For instance, FreeDA \citep{freeda} relies on predefined categories from COCO Captions \citep{cococaption}, ReME \citep{reme} utilizes Large Language Models to generate textual labels, and PGSeg \citep{UPK} depends on massive image-text pair datasets. A critical limitation of these approaches is that they are inherently bounded by a fixed set of text vocabularies during construction. This contradicts the objective of OVSS. Text-anchored memory banks force continuous visual features into predefined semantic bins, inevitably overfitting to the taxonomy and granularity of the construction dataset. In contrast, our method decouples the memory bank construction from language and naturally accommodates the open-ended nature of OVSS. This language-free design elegantly circumvents vocabulary overfitting, unleashing the true potential of memory banks for infinite-category scenarios.

\section{Method}\label{sec4} \label{sec:method}

\begin{figure*}[t]
\centering
\includegraphics[width=1\linewidth]{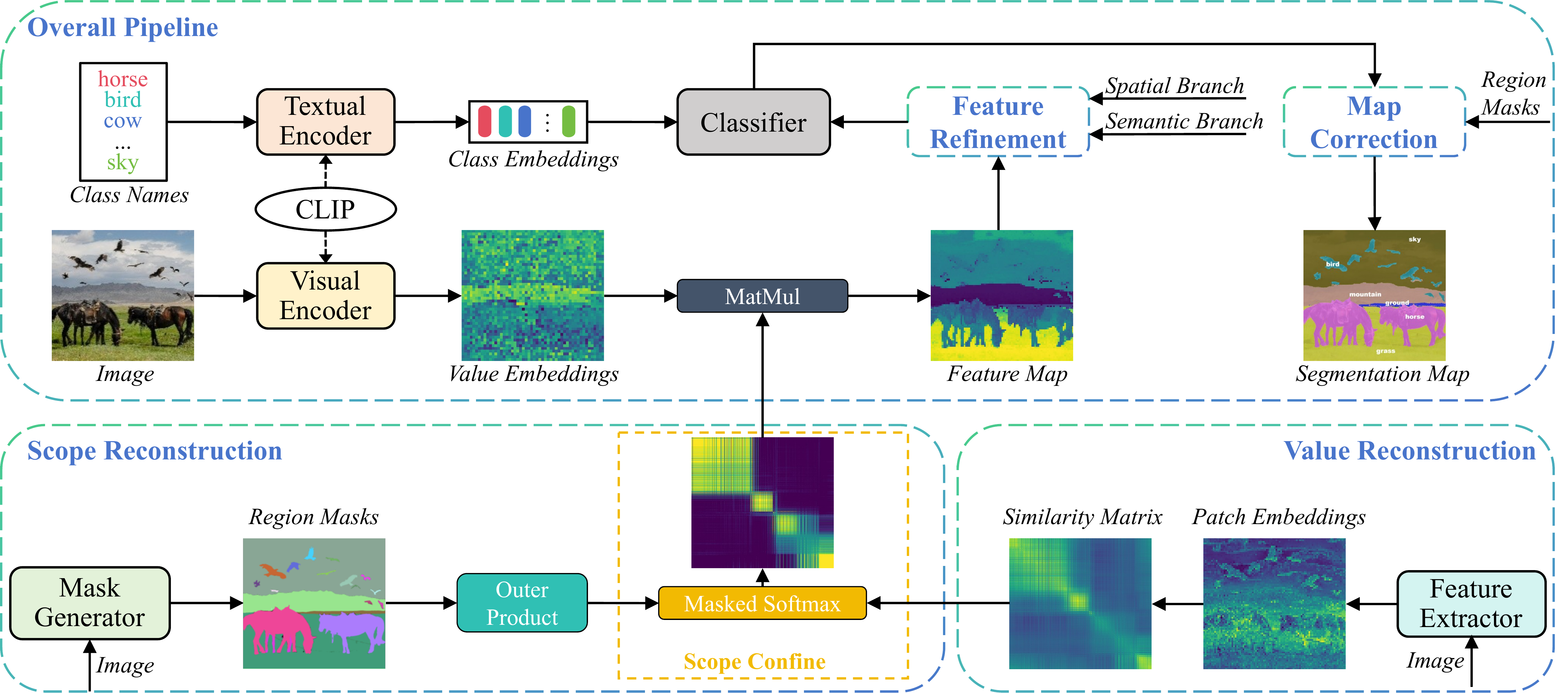}
\caption{The overall framework of CorrCLIP. CLIP’s visual encoder processes the image to obtain value embeddings of image patches, while CLIP’s textual encoder encodes class names to produce class embeddings. A mask generator creates region masks and a feature extractor derives coherent patch embeddings to compute the similarity matrix for value reconstruction. Applying the region masks to the similarity matrix achieves scope reconstruction. The feature map is further refined via spatial and semantic branches, and the refined features are compared with class embeddings to generate the segmentation map, which is finally updated using the region masks}
\label{fig_framework}
\end{figure*}

In this section, we elaborate on the proposed CorrCLIPv2. We begin by establishing the necessary preliminaries in \Cref{Preliminary}. Subsequently, \Cref{corrclipv1} revisits the core architecture of the conference version, CorrCLIP, detailing its four integral modules: Scope Reconstruction, Value Reconstruction, Feature Refinement, and Map Correction. The overall framework is shown in \Cref{fig_framework}. Building upon this framework, \Cref{corrclipv2} introduces the advanced extensions proposed in this journal version, which are designed to mitigate contextual bias through cross-image retrieval-augmented mechanisms.

\subsection{Preliminary} \label{Preliminary}
Our method leverages the pre-trained ViT visual encoder from CLIP. The input image is initially serialized into a token sequence $\bm{X}_C \in \mathbb{R}^{N \times d}$ augmented with positional encodings, where $N$ and $d$ denote the sequence length and embedding dimension, respectively. Prior to the final layer, the sequence propagates through stacked standard transformer blocks, each processing features via multi-head self-attention and feed-forward networks with residual connections.

In this work, our modifications primarily focus on the final layer of the encoder to tailor the features for segmentation tasks. Within this last layer, the input token sequence $\bm{X}_C$ is mapped to generate the query, key, and value embeddings, formally denoted as $\bm{Q}_C, \bm{K}_C$, and $\bm{V}_C \in \mathbb{R}^{N \times d}$, respectively. For brevity and clarity of notation, we formulate the subsequent operations based on single-head attention. 
Departing from the standard attention mechanism, the similarity matrix $\bm{S} \in \mathbb{R}^{N \times N}$ is calculated by the dot product between query embeddings themselves, rather than between queries and keys. As demonstrated in ClearCLIP~\citep{lan2024clearclip}, this self-query correlation significantly improves coherence:
\begin{equation}
    \bm{S} = \bm{Q}_C\bm{Q}_C^\mathrm{T}, \label{similarity}
\end{equation}
where $\bm{S}$ explicitly represents the patch-to-patch correlations, capturing the semantic affinity across different patches of the image. The computation of $\bm{S}$ is pivotal, as it serves as the foundation for enhancing CLIP's dense prediction capabilities. In the subsequent subsections, we will elaborate on our methodology to reconstruct these patch correlations for improving segmentation performance.

To derive the final attention weights, a row-wise softmax operation is applied to the scaled similarity matrix. These weights are then used to aggregate the semantic information from $\bm{V}_C$. Following the insights from ClearCLIP~\citep{lan2024clearclip}, we remove both the residual connections and the feed-forward network in this final layer to reduce the noise and yield cleaner segmentation results. Consequently, the value embeddings aggregated by the attention map yield the final image representations $\bm{F}_{img} \in \mathbb{R}^{N \times d}$:
\begin{align}
    &\bm{A} = \operatorname{Softmax}\left(\frac{\bm{S}}{\sqrt{d}}\right) \label{attn} ,\\ 
    &\bm{F}_{img} = \bm{A} \bm{V}_C.
    \label{attn_value}
\end{align}

In parallel with the visual stream, the textual branch processes semantic category information. We construct category descriptions by combining $C$ distinct class names with standard ImageNet prompt templates \citep{clip}. These descriptions are encoded by CLIP's textual encoder to yield the class embeddings $\bm{F}_{txt} \in \mathbb{R}^{C \times d}$. To enable zero-shot segmentation, the image patch features $\bm{F}_{img}$ are projected into a common feature space aligned with $\bm{F}_{txt}$. Finally, the segmentation logits $\bm{P} \in \mathbb{R}^{N\times C}$ are generated by treating the class embeddings as the weights of a linear classifier:
\begin{align}
    \bm{P} = \bm{F}_{img} \bm{F}_{txt}^\mathrm{T}.
\end{align}
The final semantic labels are obtained via an argmax operation on $\bm{P}$.

\subsection{CorrCLIPv1: Correlation Reconstruction in Single Image} \label{corrclipv1}
\subsubsection{\textbf{Scope Reconstruction}} \label{Scope}
Although the self-query attention mechanism improves segmentation performance, it remains constrained by the intrinsic incoherence of CLIP's patch embeddings, which inevitably leads to undesirable inter-class correlations. To address this, we propose Scope Reconstruction, which explicitly reconstructs the interaction scope to disentangle inter-class propagation.

Specifically, we leverage a class-agnostic mask generator to provide structural priors. The generator produces a set of binary region masks $\bm{M} = [\bm{m}_1, \bm{m}_2, ..., \bm{m}_Z] \in \mathbb{R}^{Z \times N}$, where $\bm{m}_i \in \{0, 1\}^N$ denotes the $i$-th region mask and $Z$ is the number of masks. The union of pixels not covered by any specific region mask forms the background mask $\bm{m}_0$. Unlike the previous conference version using complex clustering to merge masks, we directly utilize these raw region masks to define the scope of patch interactions, which avoids the computational overhead of clustering.

To define the interaction scope, we construct a binary matrix $\bm{E} \in \mathbb{R}^{N \times N}$, where $\bm{E}_{i,j}=1$ allows attention between patch $i$ and $j$, and $\bm{E}_{i,j}=0$ inhibits it.
For patches falling within a specific region $\bm{m}_k$ ($k > 0$), interactions are strictly confined to that region. For the background region $\bm{m}_0$, where semantic boundaries are less defined, we apply a dynamic thresholding strategy based on the global mean of the similarity matrix to filter out weak connections. Formally, the interaction matrix is defined as:
\begin{equation}
    \bm{E} = \mathbb{I}\!\left(\bm{S} > \bar{\bm{S}}\right) \odot (\bm{m}_0 \otimes \bm{m}_0) + \sum_{k=1}^Z \bm{m}_k \otimes \bm{m}_k,
    \label{semantic-matrix}
\end{equation}
where $\otimes$ denotes the outer product, $\odot$ is the element-wise product, $\bar{\bm{S}}$ is the mean of $\bm{S}$, and $\mathbb{I}(\cdot)$ is the indicator function.
Consequently, the standard softmax in \Cref{attn} is replaced by a Masked Softmax, which enforces the structural constraints defined by $\bm{E}$:
\begin{equation}
    \bm{A} = \operatorname{Masked~Softmax}\left(\frac{\bm{S}}{\sqrt{d}},~\bm{E}\right). \label{masked attention}
\end{equation}
This operation effectively zeros out attention scores where $\bm{E}_{i,j}=0$, ensuring that feature aggregation occurs exclusively within coherent regions.

\subsubsection{\textbf{Value Reconstruction}} \label{Value} 
While Scope Reconstruction explicitly reduces inter-class patch correlations, it relies on the quality of the generated masks. A critical issue arises when the mask generator produces inaccurate masks which encompass multiple distinct semantic categories. In such cases, the binary constraint of Scope Reconstruction fails to prevent correlations between these distinct classes.

To mitigate the adverse effects of such mask inaccuracies, we introduce Value Reconstruction. The objective is to refine the internal affinity measure within the permitted scope, ensuring that patches only aggregate information from semantically consistent neighbors when the spatial mask is imperfect. 
SSL-based VFMs possess a more intrinsic understanding of fine-grained semantic layouts compared to CLIP, whose visual features lack local discriminability.
Therefore, we leverage a VFM to reconstruct the similarity matrix. Let $\bm{F}_{{vfm}} \in \mathbb{R}^{N \times d'}$ denote the feature sequence from the final layer of the VFM. Unlike the conference version, which relies on DINO's internal query and key embeddings, we instead employ the final output features. This modification ensures broader applicability, as the exceptional semantic alignment inherent in DINO's internal spaces does not necessarily generalize to other VFMs. The reconstructed similarity matrix $\bm{S}$ is computed as:
\begin{equation}
    \bm{S} = \frac{\bm{F}_{vfm} \bm{F}_{vfm}^\mathrm{T}}{\Vert \bm{F}_{vfm} \Vert^2}.
\end{equation}

This VFM-derived similarity acts as a soft semantic filter. Even if a mask erroneously permits interaction between two different objects, the high discriminative power of the VFM features ensures that their similarity in $\bm{S}$ remains low. Consequently, the contribution of these irrelevant patches is suppressed during aggregation. The final attention mechanism combines the hard structural constraint of the mask $\bm{E}$ with the soft semantic guidance of $\bm{S}$:
\begin{equation}
    \bm{A} = \operatorname{Masked~Softmax}\left(\frac{\bm{S}}{\tau},~\bm{E}\right),
\end{equation}
where $\tau < 1$ is a temperature coefficient used to sharpen the attention distribution.

\subsubsection{\textbf{Feature Refinement}} \label{Refinement}
With the correlation reconstruction, we obtain the aggregated image features. To further bolster the discriminative power of these features, we introduce a dual-branch Feature Refinement strategy comprising a Semantic Branch and a Spatial Branch.

The class token in CLIP aggregates information across the entire image, which dilutes region-specific semantics. To enrich the region-specific semantic information, we introduce a Semantic Branch incorporating Mask Class Tokens (MCT). Specifically, we append $Z$ class tokens $\bm{T}_{mask} \in \mathbb{R}^{Z \times d}$ to the input sequence, corresponding to the $Z$ regions. During the forward pass, these tokens exclusively attend to image tokens within their assigned masks, effectively aggregating a global representation for each region. Subsequently, we obtain the dense feature map by redistributing these region-level representations back to patch-level:
\begin{equation}
    \bm{X}_{MCT} = \bm{M}^\mathrm{T} \bm{T}_{mask}.
\end{equation}

Additionally, motivated by recent findings~\citep{sc-clip, resclip, cliper} indicating that the lower layers of CLIP preserve superior spatial coherence compared to the final layer, we incorporate these lower layer features to restore spatial details. Specifically, to align these features with the final feature space, we project the token sequence from the preceding layers using the value projection of the final layer. This operation results in spatially discriminative value embeddings $\bm{V}_C'$.

Finally, image features $\bm{F}_{img}$ are updated as a weighted fusion of the main branch, the spatial branch, and the semantic branch:
\begin{equation}
    \bm{F}_{img} = \bm{A} \bm{V}_C + \alpha \bm{A} \bm{V}_C' + \beta \bm{X}_{MCT}, 
\end{equation}
where $\alpha$ and $\beta$ are hyperparameters balancing the contributions of spatial details and semantic context, empirically set to $1.0$ and $0.5$, respectively.

\subsubsection{\textbf{Map Correction}} \label{Correction}
Pixel-wise predictions from CLIP can exhibit high variance, resulting in noise or inconsistent labels within a single object. 
Since our Scope Reconstruction already leverages mask priors, we extend this logic to the output space. We assume that pixels belonging to the same generic mask likely belong to the same semantic category. Therefore, we perform a majority voting operation within each mask region $\bm{m}_k$ ($k > 0$) to unify the predictions:
\begin{equation}
    \bm{P}[\bm{m}_k] \leftarrow \operatorname{Mode}( \bm{P}[\bm{m}_k]).
\end{equation}
Note that in this context, $\bm{P}$ denotes the predicted segmentation map. This map correction step effectively regularizes the prediction using geometric priors, smoothing out outliers and ensuring spatially consistent segmentation results.

\subsection{CorrCLIPv2: Cross-Image Retrieval-Augmented Paradigm} \label{corrclipv2}
\subsubsection{\textbf{Empirical Analysis and Design Rationale}} \label{VISR}
While CorrCLIP effectively mitigates inter-class interference in the final attention layer, the preceding ViT blocks in CLIP inherently rely on global interactions. Consequently, the feature representation of a local object is inevitably entangled with its surrounding context. This dependency causes objects to exhibit semantic drift when appearing in complex or ambiguous backgrounds, leading to misclassification.
We hypothesize that while individual object representations are susceptible to specific contextual bias, the semantic prototype distilled from multiple instances of the same category is robust. Theoretically, aggregating features from diverse contexts can marginalize out context-induced variance, thereby stabilizing the semantic representation and prediction. To validate this hypothesis, we conduct a controlled \textit{Oracle Study}.

\begin{figure}[t]
\centering
\includegraphics[width=1\linewidth]{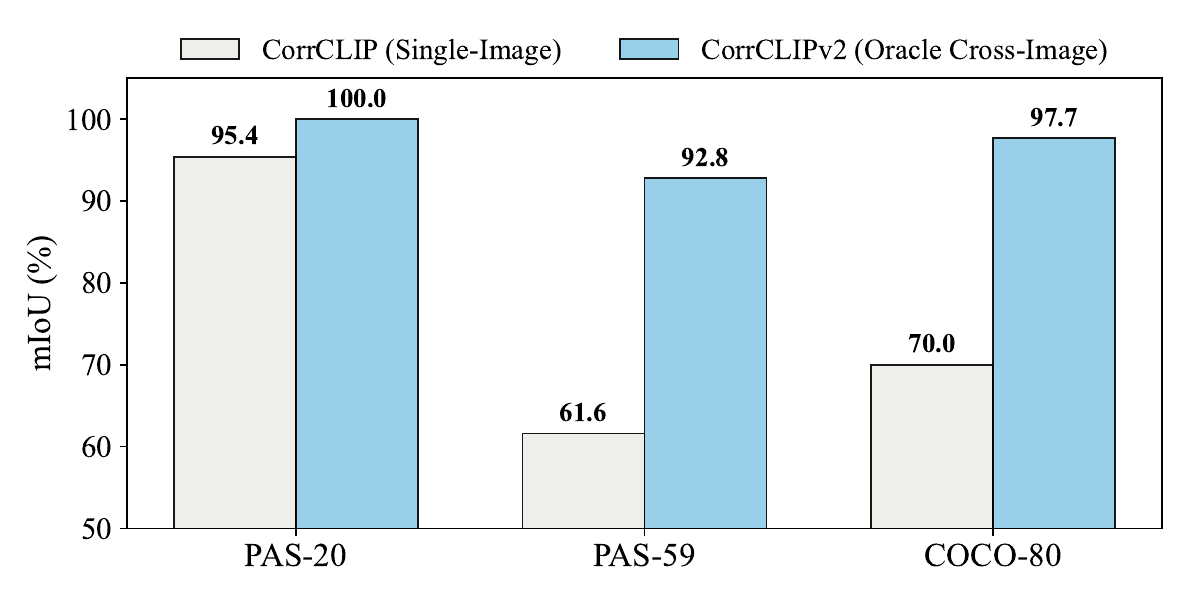}
\caption{Oracle study on cross-image semantic aggregation. We compare the single-image CorrCLIP baseline against an Oracle CorrCLIPv2 variant that aggregates features from a ground-truth support set. 
The substantial performance leap, particularly on challenging datasets (e.g., +31.2\% on PAS-59 and +27.7\% on COCO-80), confirms that semantic prototype distilled from multiple instances effectively mitigates context-induced interference
}
\label{fig_ideal_bank}
\end{figure}

\paragraph{Oracle Validation}
We construct an ideal feature bank using ground-truth class masks. Specifically, we extract CorrCLIP features for all objects in a dataset and group them by category. During inference, for each target object, we randomly sample a support set from the corresponding category bank and average them with the target feature. As illustrated in \Cref{fig_ideal_bank}, this oracle aggregation yields a substantial performance improvement over the single-image baseline. This result empirically confirms that accessing a semantically consistent support set can effectively mitigate feature context-induced interference. Motivated by this, CorrCLIPv2 aims to approximate this oracle behavior in an unsupervised, open-vocabulary setting via retrieval.

\begin{figure}[t]
\centering
\includegraphics[width=1\linewidth]{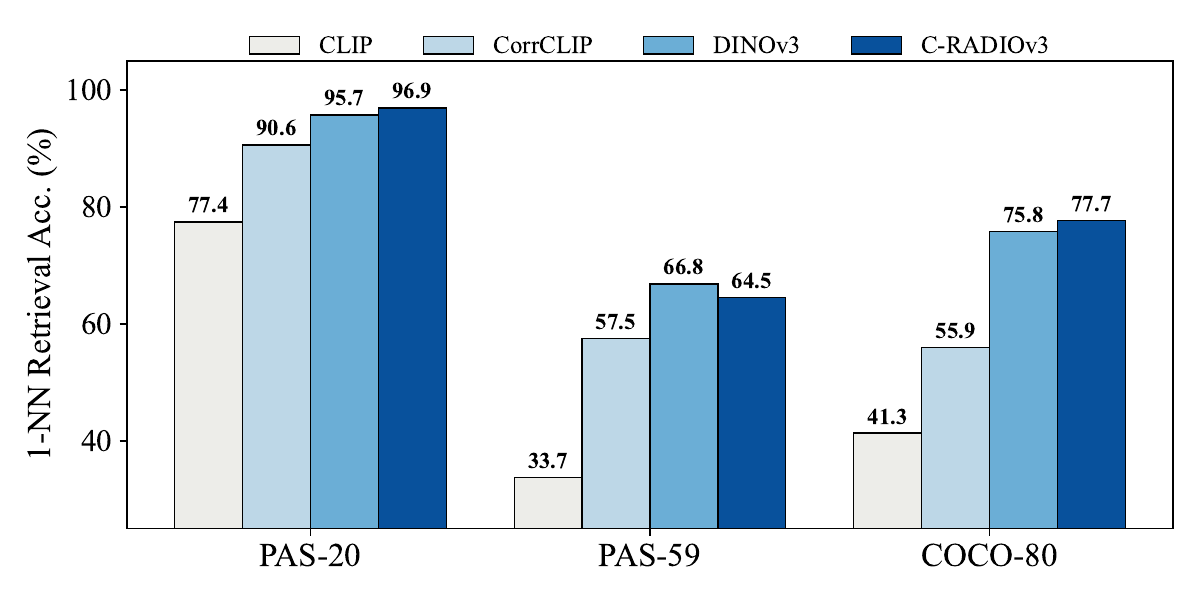}
\caption{Evaluation of feature discriminability of different models. We report the 1-NN retrieval accuracy to quantify how well different models maintain semantic consistency. While CorrCLIP improves upon the original CLIP, SSL-based VFMs exhibit significantly higher accuracy across all benchmarks. This observation justifies our design choice of utilizing VFMs as the index to ensure robust and accurate support set retrieval}
\label{fig_ideal_acc}
\end{figure}

\paragraph{VFM-Based Indexing Strategy}
The efficacy of the retrieval mechanism hinges on the quality of the index. An optimal index should ensure that the nearest neighbors of a query object belong to the same semantic category, independent of background variations. To identify the most suitable feature space for indexing, we conduct a 1-Nearest Neighbor (1-NN) retrieval consistency analysis.
We extract features for all ground-truth objects using four models: the original CLIP, our CorrCLIP, DINOv3 \citep{dinov3} and C-RADIOv3 \citep{radiov2}. We then measure the accuracy that an object's nearest neighbor shares the same class label.
As shown in \Cref{fig_ideal_acc}, original CLIP exhibits poor retrieval consistency due to its lack of local discriminability. While CorrCLIP promotes consistency by reconstructing patch correlations, it still falls short of SSL-based VFMs, which demonstrate superior discriminability, achieving near-perfect consistency on VOC \citep{pascal2012} and significant leads on complex datasets like COCO \citep{coco}.

\begin{figure}[t]
\centering
\includegraphics[width=1\linewidth]{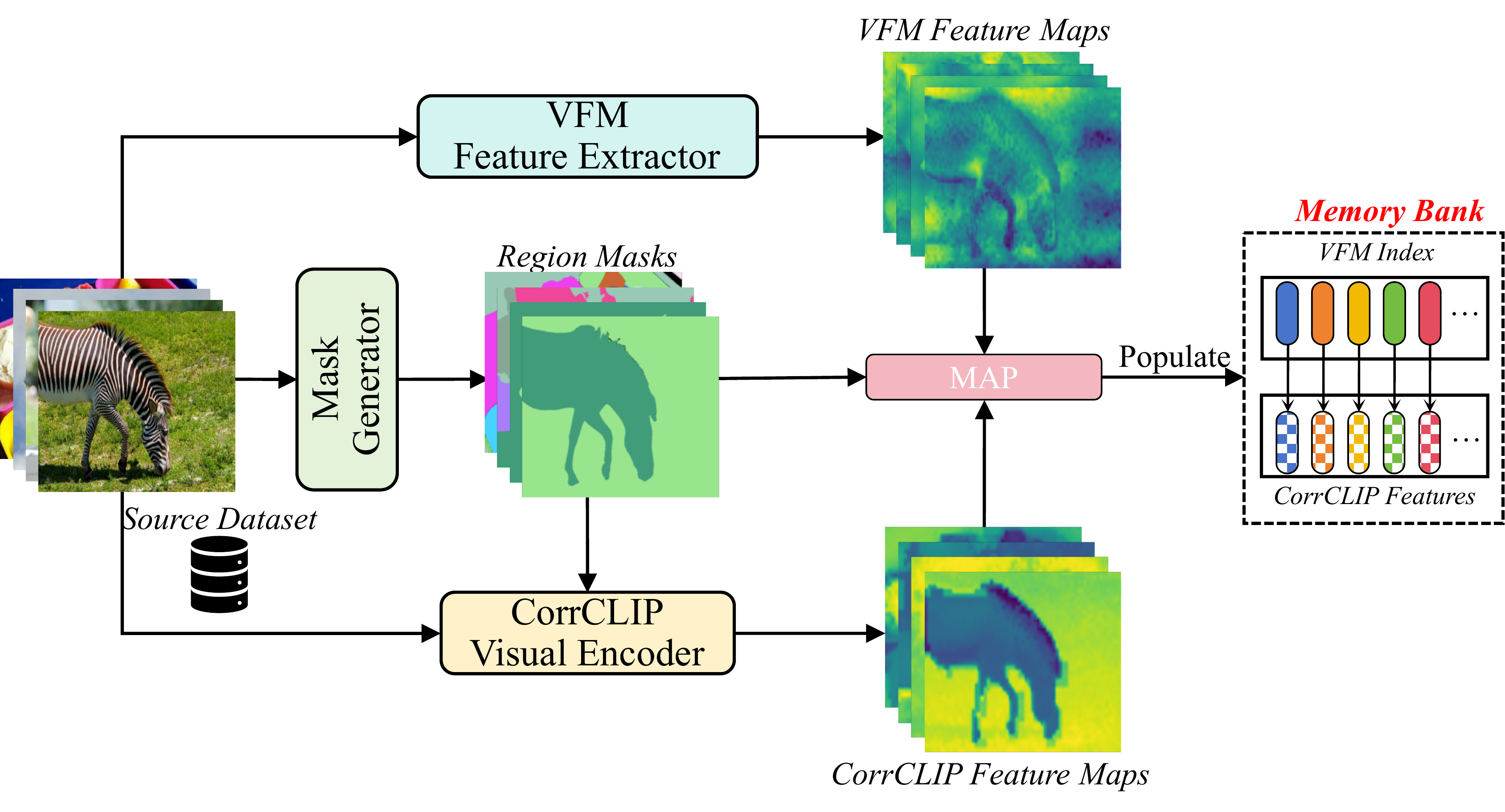}
\caption{Construction of the offline feature memory bank. For each image in the unlabeled collection, we extract object embeddings by applying Mask Average Pooling (MAP) to both VFM and CorrCLIP features across the generated region masks. The resulting embeddings populate the bank where VFM features as indexing keys and CorrCLIP features as semantic values }
\label{fig_bank_construction}
\end{figure}

\subsubsection{\textbf{VFM-Indexed Semantic Retrieval}} \label{VISR_method}

Based on the aforementioned insights, we utilize SSL-based VFMs as the indexing keys to accurately locate the support set, while storing the open-vocabulary CorrCLIP features as the semantic values. This strategy effectively combines the fine-grained discriminability of VFMs with the text-alignment capabilities of CLIP.
\paragraph{Offline Memory Bank Construction}
As illustrated in \Cref{fig_bank_construction}, we first build a memory bank using a large-scale unlabeled dataset $\mathcal{D} = \{I_i\}_{i=1}^{|\mathcal{D}|}$. For each image $I \in \mathcal{D}$, we generate a set of class-agnostic binary masks $\bm{M}_{b} = [\bm{m}^{bank}_{1}, \dots, \bm{m}^{bank}_{Z'}] \in \mathbb{R}^{Z' \times N}$ to define object proposals.
To populate the bank, each image is processed through two parallel encoding paths. First, in the index path (key generation), we employ the same VFM encoder used in the Value Reconstruction to extract a dense feature map $\bm{F}^{bank}_{vfm} \in \mathbb{R}^{N \times d'}$. Simultaneously, in the semantic path (value generation), the image is processed by our CorrCLIP visual encoder to yield the semantic feature map $\bm{F}^{bank}_{clip} \in \mathbb{R}^{N \times d}$.
To construct object-centric entries for the bank, we aggregate these dense features within each mask region using Mask Average Pooling (MAP). For the $j$-th object mask $\bm{m}^{bank}_{j}$, the indexing key embedding $\bm{k}_j$ and the semantic value embedding $\bm{v}_j$ are derived as:
\begin{equation}
\begin{aligned}
    \bm{k}_j &= \operatorname{MAP}(\bm{F}^{bank}_{vfm}, \bm{m}^{bank}_{j}), \\
    \bm{v}_j &= \operatorname{MAP}(\bm{F}^{bank}_{clip}, \bm{m}^{bank}_{j}).
\end{aligned}
\end{equation}
By iterating over the entire dataset $\mathcal{D}$, we collect all object pairs to form the VFM index bank $\mathcal{B}_{vfm}=\{\bm{k}_n\}_{n=1}^{N_{total}}$ and the CorrCLIP feature bank $\mathcal{B}_{clip}=\{\bm{v}_n\}_{n=1}^{N_{total}}$, where $N_{total}$ denotes the total number of object instances. 

\begin{figure}[t]
\centering
\includegraphics[width=0.95\linewidth]{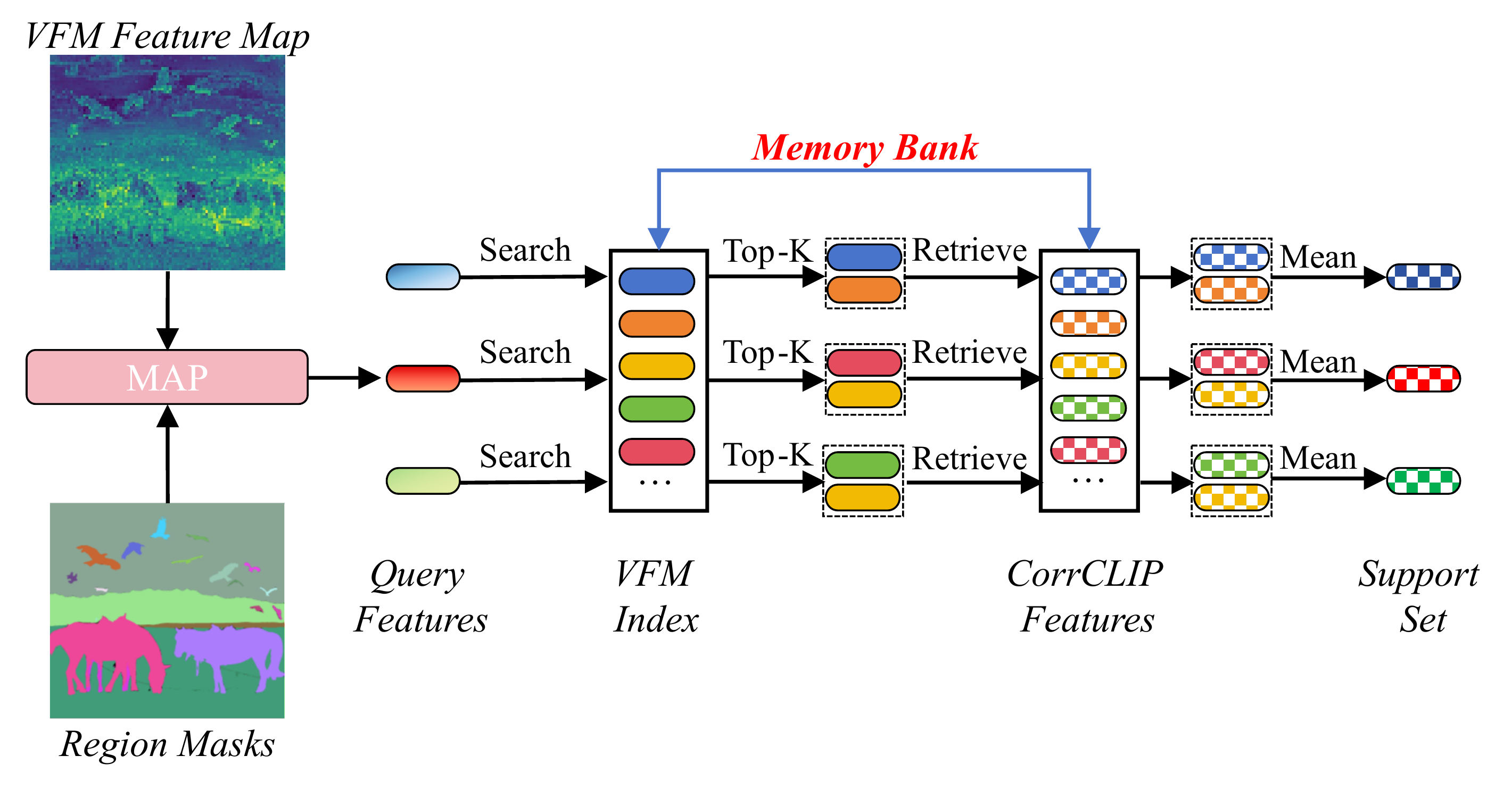}
\caption{Illustration of the retrieval process during inference. The generated region masks are applied to the VFM feature map via Mask Average Pooling (MAP) to extract object embeddings. For each query embedding, we identify the Top-$K$ nearest neighbors in the VFM bank to obtain the corresponding indices. Then the associated CorrCLIP features are retrieved and averaged to construct the support set}
\label{fig_retrieve}
\end{figure}

\paragraph{Inference and Retrieval}
During the inference phase, as depicted in \Cref{fig_retrieve}, we leverage the pre-computed memory bank to mitigate contextual bias in the target image. Crucially, this retrieval can be seamlessly integrated into CorrCLIP pipeline without any additional computations. We directly utilize the region masks $\bm{M}$ obtained from Scope Reconstruction (\Cref{Scope}) and the VFM feature map $\bm{F}_{vfm}$ generated during Value Reconstruction (\Cref{Value}). Using these pre-existing representations, we derive the query set $\mathcal{Q} = \{\bm{q}_1, \dots, \bm{q}_Z\}$ for the target image, where:
\begin{equation}
    \bm{q}_j = \operatorname{MAP}(\bm{F}_{vfm}, \bm{m}_j).
\end{equation}

For each query $\bm{q}_j$, we perform a Top-$K$ search within $\mathcal{B}_{vfm}$ to identify semantically consistent instances. To ensure scalability and efficiency given the large size of $N_{total}$, we employ Inverted File with Product Quantization (IVFPQ) for approximate nearest neighbor search:
\begin{equation}
    \mathcal{I}_j = \operatorname{IVFPQ}(\bm{q}_j, \mathcal{B}_{vfm}, K),
\end{equation}
where $\mathcal{I}_j$ represents the indices of the Top-$K$ retrieved neighbors. Subsequently, we fetch the corresponding semantic features from $\mathcal{B}_{clip}$ using these indices. The support feature $\bm{s}_j$ for the $j$-th region is computed by averaging these retrieved features:
\begin{equation}
    \bm{s}_j = \frac{1}{K} \sum_{i \in \mathcal{I}_j} \mathcal{B}_{clip}[i].
\end{equation}

Collecting the support vectors for all regions yields the support set $\bm{V}_{supp} = [\bm{s}_1, \dots, \bm{s}_{Z}]\in \mathbb{R}^{Z \times d}$. We then redistribute these region-level support features back to the dense patch-level:
\begin{equation}
    \bm{F}_{supp} = \bm{M}^\mathrm{T} \bm{V}_{supp},
\end{equation}
where $\bm{F}_{supp} \in \mathbb{R}^{N \times d}$ represents the retrieved support feature map.
Finally, the segmentation prediction is refined by fusing the original single-image logits with the retrieval-augmented support logits:
\begin{equation}
    \bm{P} = \theta (\bm{F}_{img} \bm{F}_{txt}^\mathrm{T}) + 
    (1-\theta) (\bm{F}_{supp} \bm{F}_{txt}^\mathrm{T}),
\end{equation}
where $\theta$ is a hyperparameter to balance the original inference and the semantic support retrieval.

\subsubsection{\textbf{Decoupled Multi-CLIP Ensemble}} \label{DMCE}
Ensembling diverse CLIP variants \citep{openclip, metaclip, dfnclip} enhances robustness but incurs prohibitive computational costs. Our framework addresses this by decoupling the expensive feature encoding of diverse CLIP variants from the online inference process. Since retrieval is governed by a shared VFM-based index, we can pre-populate the memory bank with CorrCLIP-enhanced features from multiple CLIP models. Consequently, auxiliary representations are simply retrieved and fused during inference, achieving the benefits of ensembling with negligible computational overhead.

Let $\bm{F}'_{supp}$ and $\bm{F}_{txt}'$ denote the retrieved support feature map and the text embeddings corresponding to the auxiliary CLIP model, respectively. The final segmentation prediction aggregates the primary model's single-image inference with the multi-source semantic support:
\begin{equation}
    \bm{P} = \theta (\bm{F}_{img} \bm{F}_{txt}^\mathrm{T}) + 
    (1 - \theta) (\frac{\bm{F}_{supp} \bm{F}_{txt}^\mathrm{T} + 
    \bm{F}'_{supp} \bm{F}_{txt}^{'\mathrm{T}}}{2}).
\end{equation}

\section{Experiments}

\subsection{Datasets and Evaluation Metrics}

We conduct comprehensive evaluations on ten standard benchmarks and a massive multi-domain benchmark.

\textbf{Standard Benchmarks.} We utilize the validation sets of five datasets, which constitute ten specific benchmarks.
\textbf{Pascal VOC} \citep{pascal2012} (1,449 images) is evaluated on V20 (20 classes) and V21 (w/ background).
\textbf{Pascal Context} \citep{context} (5,104 images) supports three benchmarks: PC59 (59 classes), PC60 (w/ background), and fine-grained PC459 (459 classes).
\textbf{COCO} \citep{coco} (5,000 images) is divided into COCO-Stuff (171 classes) and COCO-Object (81 classes).
\textbf{ADE20K} \citep{ade} (2,000 images) is evaluated on the standard A150 (150 classes) and the large-vocabulary A847 (847 classes).
\textbf{Cityscapes} \citep{cityscapes} (500 images) focuses on urban scenes with 19 classes.

\textbf{MESS Benchmark.} To verify open-vocabulary robustness across distinct distributions, we employ the Multi-Domain Evaluation of Semantic Segmentation (MESS) benchmark \citep{mess}. It comprises 22 datasets spanning five diverse domains: {General}, {Earth Monitoring}, {Medical Sciences}, {Engineering}, and {Agriculture \& Biology}.

\textbf{Memory Bank Source.} We construct our offline feature memory bank using the \textbf{COCO training set} \citep{coco}, which contains approximately 118k images. Crucially, we utilize only the raw images to extract visual features, without accessing any ground-truth annotations or semantic labels.

Following standard protocols, we report the mean Intersection over Union (mIoU) for all experiments.

\subsection{Implementation Details}

We instantiate our framework with two variants based on model capacity: CorrCLIPv2-B and CorrCLIPv2-L, which utilize the CLIP-ViT-B/16 and CLIP-ViT-L/14 architectures, respectively. Unless otherwise specified, we employ the weights from MetaCLIP \citep{metaclip} as the primary backbone. For the Decoupled Multi-CLIP Ensemble, we additionally leverage the DFN \citep{dfnclip} as the auxiliary backbone to incorporate diverse pre-training priors.
For the class-agnostic mask generator, we employ EntitySeg \citep{entityseg} implemented with the Mask2Former \citep{mask2former} architecture and a Hornet-L \citep{hornet} backbone. For VFM, we utilize C-RADIOv3 \citep{radiov2} with a ViT-L/16 backbone.

The VFM-Indexed Semantic Retrieval is implemented using the FAISS-GPU library \citep{faiss}. To balance efficiency and accuracy on large-scale databases, we utilize the IVFPQ index. 
During inference, we adopt a sliding window strategy to handle high-resolution inputs while preserving local details. The temperature coefficient $\tau$ for the similarity calculation is set to 0.25. The balancing hyperparameter $\theta$ is set to 0.5.

\begin{table*}[t]
\caption{Comparison with state-of-the-art OVSS methods on 10 standard benchmarks. We report the best performance for each method. The best and second-best training-free methods are highlighted in bold and underlined, respectively  \label{tab:comparison_main}}
\centering
 \resizebox{\textwidth}{!}{
\begin{tabular}{l|*{10}{c}|c}
    \toprule
    Method & V21 & V20 & PC60 & PC59 & Object & Stuff & A150 & City & A847 & PC459 & Mean \\

    \midrule
    \multicolumn{12}{c}{\textbf{\emph{Training-based (Fully Supervised)}}} \\
    SAN~\citep{san++} & - & 95.4 & - & 57.7 & - & - & 32.1 & - & 12.4 & 15.7 & - \\
    CLIPSelf~\citep{clipself} & - & - & - & 62.3 & - & - & 34.5 & - & 12.4 & - & - \\
    CAT-Seg~\citep{catseg} & 82.5 & 97.0 & - & 63.3 & - & - & 37.9 & - & 16.0 & 23.8 & - \\
    SED++~\citep{sed++} & - & 95.9 & - & 61.0 & - & - & 38.4 & - & 16.0 & 24.1 & - \\    

    \midrule
    \multicolumn{12}{c}{\textbf{\emph{Training-based (Weakly Supervised)}}}   \\
    CLIP-DINOiser~\citep{clip-dinoier} & 62.1 & 80.9 & 32.4 & 35.9 & 34.8 & 24.6 & 20.0 & 31.7 & 7.1 & 8.4& -\\
    dino.txt~\citep{dino.txt}  & - & 62.1 & - & 30.9 & - & 20.9 & 20.6 & 32.1 & - & - & -\\
    Talk2DINO~\citep{Talk2DINO}  & 65.8 & 88.5 & 37.7 & 42.4 & 45.1 & 30.2 & 22.5 & 38.1 & - & - & -\\
    Seg4Diff~\citep{seg4diff} & - & 89.8 & - & 51.2 & - & - & 35.2 & 26.0 & - & - & -\\

     \midrule
    \multicolumn{12}{c}{\textbf{\emph{Training-free (CLIP Only)}}}   \\

    CLIP~\citep{clip} & 11.5 & 41.9 & 4.4 & 9.2 & 1.6 & 4.4 & 2.9 & 5.0 & 0.8 & 1.3 & 8.3\\

    CaR~\citep{car} & 67.6 & 91.4 & 30.5 & 39.5 & 36.6 & 11.2 & 17.7 & 15.1 & 5.0 & 11.5 & 32.6 \\

    GEM~\citep{grounding-everything} & 46.2 & 73.0 & 33.0 & 35.1 & 33.7 & 24.2 & 15.6 & 30.8 & 6.2 & 11.2 & 30.9\\

    CLIPtrase~\citep{cliptrase} & 53.0 & 81.2 & 30.8 & 34.9 & 44.8 & 24.1 & 17.0 & 21.1 & 5.9 & 9.9 & 32.3\\

    SCLIP~\citep{sclip} & 61.7 & 83.5 & 31.5 & 36.1 & 32.1 & 23.9 & 17.8 & 34.1 & 5.9 & 6.7 & 33.3\\

    ClearCLIP~\citep{lan2024clearclip} & 51.8 & 80.9 & 32.6 & 35.9 & 33.0 & 23.9 & 16.7 & 30.0 & 6.2 & 7.8 & 31.9\\

    NACLIP~\citep{naclip} & 64.1 & 83.0 &  35.0 & 38.4 & 36.2 &  25.7 & 19.1 & 38.3 & 6.5 & 9.0 & 35.5\\

    CDAM~\citep{CDAM} & 58.7 & 75.2 & 30.6 & 35.5 & 35.2 & 24.8 & 17.2 & 24.8 & 6.4 & 7.3 & 31.6\\

    ResCLIP~\citep{resclip}  & 61.1 & 86.0 & 33.5 & 36.8 & 35.0 & 24.7 & 18.0 & 35.9 & 6.7 & 7.7 & 34.5\\

    SFP~\citep{sfp} & 63.9 & 84.5 & 37.2 & 39.9 & 37.9 & 26.4 & 20.8 & 41.1 & 7.5 & 8.0 & 36.7\\

    FreeCP~\citep{FreeCP} & 64.5 & 81.5 & 35.7 & 39.3 & 36.9 & 26.1 & 18.9 & 34.4 & 6.7 & 8.2 & 35.2\\

    SC-CLIP~\citep{sc-clip} & 65.0 & 88.3 & 36.9 & 40.6 & 40.5 & 26.9 & 21.7 & 41.3 & 9.3 & 9.7 & 38.0\\

    \midrule
    \multicolumn{12}{c}{\textbf{\emph{Training-free (VFM Based)}}}   \\

    PnP-OVSS~\citep{PnP-OVSS}& 51.3 & 79.1 & 28.0 & 31.0 & 36.2 & 17.9 & 14.2 & 19.3 & 4.2 & 5.5 & 28.7\\

    FreeDA~\citep{freeda} & 55.4 & 87.9 & 38.3 & 43.5 & 37.4 & 28.8 & 23.2 & 36.7 & 5.3 & 10.2 & 36.7\\

    OVDiff~\citep{OVDiff}& 68.4 & 80.9 & 31.2 & 32.9 & 36.2 & 20.3 & 14.1 & 23.4 & 6.6 & 12.0 & 32.6\\

    ProxyCLIP~\citep{proxyclip}& 65.0 & 83.3 & 35.4 & 39.6 & 38.6 & 26.8 & 24.2 & 42.0 & 11.1 & 9.8 & 37.6\\

    LPOSS+~\citep{lposs}  & 62.4 & 79.3 & 35.4 & 38.6 & 34.3 & 26.5 & 22.3 & 37.9 & 8.1 & 9.5 & 35.4\\

    CASS~\citep{cass}  & 65.8 & 87.8 & 36.7 & 40.2 & 37.8 & 26.7 & 20.4 & 39.4 & 7.6 & 8.4 & 37.1\\

    FSA~\citep{FSA} & 67.9 & 85.7 & 36.3 & 40.5 & 40.2 & 27.3 & 24.5 & 43.6 & 11.2 & 10.0 & 38.7\\

    Trident~\citep{trident} & 70.8 & 88.7 & 40.1 & 44.3 & 42.2 & 28.6 & 26.7 & 47.6 & 11.1& 10.3 & 41.0\\

    CLIPer~\citep{cliper} & 72.2 & 89.8 & 39.5 & 44.6 & 44.7 & 30.4 & 25.0 & 42.5 & 9.4 & 14.7 & 41.3\\

    ReME~\citep{reme} & \underline{82.2} & \underline{93.2} & 44.6 & \underline{53.1} & 48.2 & 33.2 & 28.2 & \underline{59.0} & 8.8 & \underline{15.8} & \underline{46.6}\\

    TestRegion~\citep{textregion}& 73.1 & 89.5 & 41.2 & 46.1 & 40.6 & 31.2 & 27.3 & 47.0 & 10.8 & 9.5 & 41.6\\

    \rowcolor{mygray}
    \textbf{CorrCLIP}(Ours) & 76.7 & 91.5 & \underline{44.9} & 50.8 & \underline{49.4} & \underline{34.0} & \underline{30.7} & 51.1 & \underline{13.5} & 14.4 & 45.7\\

    \rowcolor{mygray}
    \textbf{CorrCLIPv2} (Ours) & \textbf{82.3} & \textbf{93.6} & \textbf{51.1} & \textbf{56.8} & \textbf{58.0} & \textbf{39.1} & \textbf{36.5} & \textbf{59.2} & \textbf{14.9} & \textbf{18.6} & \textbf{51.0}\\
    \bottomrule
  \end{tabular}
 }
\end{table*}

\begin{table*}[t]
\caption{Comparison with state-of-the-art OVSS methods on the 22 datasets of the MESS benchmark. All methods use their base versions. The best methods are highlighted in bold
\label{tab:comparison_mess}}
 \resizebox{\textwidth}{!}{
\begin{tabular}{l|rrrrrr|rrrrr|rrrr|rrrr|rrr|r}
\toprule

 & \multicolumn{6}{c|}{General} & \multicolumn{5}{c|}{Earth Monitoring} & \multicolumn{4}{c|}{Medical Sciences} & \multicolumn{4}{c|}{Engineering} & \multicolumn{3}{c|}{Agri. and Bio.} & \\
 & \rotatebox[origin=l]{90}{BDD~\citep{yu2020bdd100k}} 
& \rotatebox[origin=l]{90}{D. Zurich~\citep{sakaridis2019guided}} 
& \rotatebox[origin=l]{90}{MHP~\citep{li2017multiple}} 
& \rotatebox[origin=l]{90}{FoodSeg~\citep{wu2021large}} 
& \rotatebox[origin=l]{90}{ATLANTIS~\citep{erfani2022atlantis}} 
& \rotatebox[origin=l]{90}{DRAM~\citep{cohen2022semantic}} 
& \rotatebox[origin=l]{90}{iSAID~\citep{waqas2019isaid}} 
& \rotatebox[origin=l]{90}{Potsdam~\citep{DatasetISPRSPotsdam}} 
& \rotatebox[origin=l]{90}{W. Floods~\citep{mateo2021towards}} 
& \rotatebox[origin=l]{90}{FloodNet~\citep{rahnemoonfar2021floodnet}} 
& \rotatebox[origin=l]{90}{UAVid~\citep{lyu2020uavid}} 
& \rotatebox[origin=l]{90}{Kvasir-I~\citep{jha2021kvasir}} 
& \rotatebox[origin=l]{90}{CHASE~\citep{fraz2012ensemble}} 
& \rotatebox[origin=l]{90}{CryoNu.~\citep{mahbod2021cryonuseg}} 
& \rotatebox[origin=l]{90}{PAXRay~\citep{seibold2022detailed}} 
& \rotatebox[origin=l]{90}{Corrosion~\citep{bianchi2021corrosion}} 
& \rotatebox[origin=l]{90}{DeepCrack~\citep{liu2019deepcrack}} 
& \rotatebox[origin=l]{90}{PST~\citep{shivakumar2020pst900}} 
& \rotatebox[origin=l]{90}{ZW-f~\citep{bashkirova2022zerowaste}} 
& \rotatebox[origin=l]{90}{SUIM~\citep{islam2020semantic}} 
& \rotatebox[origin=l]{90}{CUB~\citep{welinder2010caltech}} 
& \rotatebox[origin=l]{90}{CWFID~\citep{haug2015crop}} 
& \rotatebox[origin=l]{90}{Mean} \\

\midrule
\multicolumn{24}{c}{\textbf{\emph{Training-based}}}  \\

X-Decoder~\citep{x-decoder} & {48} & 25 & 4 & 3 & 28 & 27 & 3 & 32 & 27 & 9 & 26 & 56 & 11 & 12 & 16 & 2 & 25 & 20 & 16 & 25 & 1 & 30 & 20 \\

SAN~\citep{san} & 38 & 25 & 9 & {20} & {37} & 50 & 5 & {38} & 32 & {38} & {42} & {70} & 18 & 12 & 20 & 4 & {51} & 20 & {22} & 23 & {17} & 6 & 27 \\

OVSeg~\citep{ovseg} & {46} & 23 & 7 & 17 & 34 & 54 & 9 & 31 & 32 & 36 & 39 & {72} & {22} & {14} & 23 & 7 & 17 & {22} & 12 & 39 & 14 & 34 & 27 \\

OpenSeeD~\citep{openseed} & {48} & {29} & 3 & 9 & 19 & 30 & 2 & 32 & 31 & 24 & 40 & 60 & {47} & {34} & 38 & 14 & 48 & 3 & 3 & 20 & 1 & 12 & 25 \\

Gr.-SAM~\citep{groundedsam} & 42 & 21 & {30} & 11 & 18 & {58} & {13} & 27 & {34} & 20 & 39 & 47 & 24 & {\textbf{39}} & 42 & {21} & {59} & {22} & 17 & 15 & 1 & {39} & {29} \\

CAT-Seg~\citep{catseg} & 47 &{29} &{24} &{27} &{41} &{66} &{20} &{46} &{36} &{38} &{42} &49 &17 &16 &32 &13 &32 &{20} &{18} &{45} &11 &{\textbf{43}} & {32} \\

DEDOS~\citep{dedos} & {49} & \textbf{34} & 29 & 31 & \textbf{45}& 70 & 21 & {47} & 41 & {41} & 42 & 64 & 25 & 32 & 45 & 14 &30 &22 &27 &{49} &\textbf{18} &38 & 37\\

\midrule
 \multicolumn{24}{c}{\textbf{\emph{Training-free}}}  \\

ClearCLIP~\citep{lan2024clearclip} & 32 & 16 & 27 & 24 & 26 & 50 & 18 & 31 & 37 & 33 & 34 & 59 & 34 & 20 & 45 & \textbf{22} & 62 & 21 & 23 & 29 & 10 & 42 & 32 \\

ProxyCLIP~\citep{proxyclip} & 38 & 20 & 32 & {30}& 32 & 58 & {23} & 37 & 42 & 36 & 38 & 77 & 28 & 17 & \textbf{48} & 21 & 69 & 22 & 26 & 40 & 9 & 42 & 36 \\

Trident~\citep{trident} & 39 & 18 & 30 & \textbf{31} & 33 & 4 & 23 & 35 & {43} & 38 & 38 & {77} & 24 & 15 & 42 & {21} & {69} & 23 & 25 & 43 & 12 & 40 & 33 \\

\rowcolor{mygray}
\textbf{CorrCLIP} (ours) & 48 & 23 & {37} & 24 & 38 & {73} & 22 & 42 & 38 & 40 & {43} & \textbf{79} & 22 & 15 & 39 & 21 & \textbf{73} & {23} & {27} & 46 & 12 & 43 & 38 \\

\rowcolor{mygray}
\textbf{CorrCLIPv2} (ours) & \textbf{54} & {32} & \textbf{40} & 27 & {41} & \textbf{83} & \textbf{24} & \textbf{49} & \textbf{46} & \textbf{43} & \textbf{44} & 74 & \textbf{47} & 39 & 43 & 21 & 70 & \textbf{25} & \textbf{28} & \textbf{56} & 6 & {43} & \textbf{43} \\

\bottomrule
\end{tabular}
    }
\end{table*}

\begin{figure*}[t]
\centering
\includegraphics[width=1\linewidth]{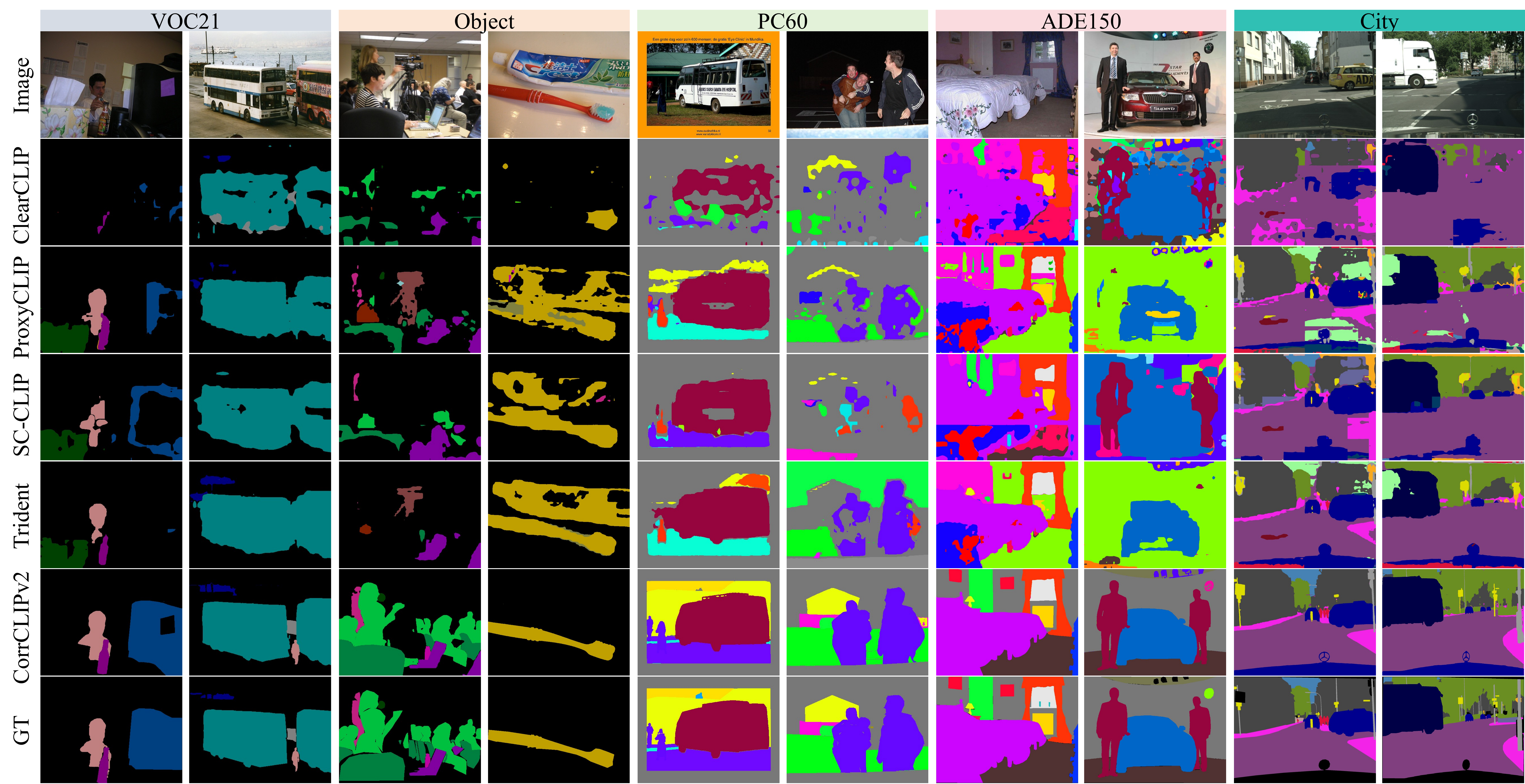}
\caption{Qualitative comparison with state-of-the-art training-free methods. While competitors often exhibit fragmentation or misclassifications, CorrCLIPv2 generates spatially coherent and semantically accurate segmentation maps, closely aligning with the Ground Truth (GT)}
\label{fig_seg_map}
\end{figure*}

\subsection{Comparison With State-of-the-Art Methods}

\subsubsection{\textbf{Comparison on Ten Standard Benchmarks}}
\Cref{tab:comparison_main} presents a comprehensive comparison between our proposed framework and state-of-the-art OVSS methods across 10 standard benchmarks. We provide a detailed analysis to elucidate the structural superiority of our approach.

Among the training-free methods, those relying solely on CLIP's internal attention modifications exhibit significantly lower performance compared to VFM-assisted methods, indicating that CLIP's intrinsic incoherence cannot be resolved by self-correction alone. More importantly, while our base model, CorrCLIP, achieves an average mIoU of 45.7\%, it significantly outperforms other multi-model synthesis approaches such as Trident \citep{trident} (41.0\% mIoU) and TestRegion \citep{textregion} (41.6\% mIoU), which utilize combinations of CLIP, SAM, and DINO. This substantial margin demonstrates that our performance gains stem from a superior architectural design—specifically, the explicit correlation reconstruction—rather than the mere brute-force aggregation of powerful pre-trained models.

We note that some recent training-free methods, such as FreeDA \citep{freeda} (36.7\% mIoU) and ReME \citep{reme} (46.6\% mIoU), also employ offline memory or prototype bank mechanisms. However, these methods tightly anchor their memory construction to language priors (e.g., FreeDA uses COCO captions, and ReME relies on Large Language Models). Our approach, CorrCLIPv2 (51.0\% mIoU), significantly surpasses them by decoupling the memory bank from text. This demonstrates that our novel language-free design avoids vocabulary overfitting and is fundamentally more suited for open-vocabulary scenarios. The superiority is particularly evident on datasets with large semantic shifts from the source dataset, such as A150 and A847, where CorrCLIPv2 outperforms ReME by 8.3\% and 6.1\% in mIoU, respectively.

To thoroughly justify our memory mechanism and prove its efficiency, we benchmark against state-of-the-art fully supervised methods (e.g., SAN  \citep{san++}, CAT-Seg \citep{catseg}, CLIPSelf \citep{clipself}, SED++ \citep{sed++}). Although our memory bank utilizes the same 118k images from the COCO dataset, it operates in a purely unsupervised, forward-pass manner without any ground-truth mask annotations or backpropagation. Remarkably, CorrCLIPv2 drastically narrows the performance gap with fully supervised models. On out-of-distribution datasets with large vocabularies (A150 and A847), CorrCLIPv2 even outperforms fully supervised methods like SAN and CLIPSelf. This validates that conventional training-based models inevitably suffer from overfitting to their predefined taxonomies, whereas our memory-augmented, training-free framework effectively preserves and scales the core visual representations of foundation models.

We also compare our method against weakly supervised approaches that attempt to distill or fine-tune VFMs using external image or text guidance, such as CLIP-DINOiser \citep{clip-dinoier}, dino.txt \citep{dino.txt}, Talk2DINO \citep{Talk2DINO}, and the diffusion-based Seg4Diff \citep{seg4diff}. CorrCLIPv2 significantly outpaces these methods across all metrics. This empirical evidence addresses the question of why a unified architecture like dino.txt falls short. It shows that dynamically reconstructing patch correlations and fusing complementary multi-model priors in a training-free framework provides a more robust, scalable, and computationally efficient solution for OVSS than weakly tuning a single foundation model.

\subsubsection{\textbf{Comparison on MESS Benchmark}}
To rigorously evaluate generalizability across varying data distributions, we extend our assessment to the MESS benchmark \citep{mess}, encompassing 22 datasets across five domains. The quantitative results are detailed in \Cref{tab:comparison_mess}.

Existing fully supervised methods, typically trained on the COCO-Stuff dataset, suffer from performance degradation when transferred to out-of-distribution scenarios. In contrast, our training-free framework demonstrates exceptional robustness. CorrCLIPv2 achieves an averaged mIoU of 43\%, surpassing the state-of-the-art fully supervised method DEDOS \citep{dedos} by a remarkable margin of 6\%. 
We attribute the superiority of CorrCLIPv2 over supervised methods to a fundamental divergence in knowledge representation. Supervised models, trained on the closed-set COCO-Stuff dataset with 171 categories, inevitably develop a bias towards the predefined taxonomy, leading to poor generalization on test data with significant distributional shifts. In contrast, CorrCLIPv2 operates without reliance on any predefined labels. By indexing raw visual features directly, our approach captures intrinsic visual patterns whose richness extends far beyond the granularity of manual annotations. Consequently, CorrCLIPv2 retrieves semantically consistent support features for concepts that may not be explicitly labeled in the source dataset, thereby achieving superior performance.

We observe slight performance drops on specific specialized datasets, such as Kvasir-Instrument, DeepCrack, and CUB-200. 
This is expected, as the generic imagery in the COCO memory bank lacks the specific visual instances required for these niche tasks. However, a distinct advantage of our paradigm is its data-efficient extensibility. Since CorrCLIPv2 relies solely on unlabeled images, which are readily accessible, the memory bank can be easily expanded without expensive retraining. As validated in our subsequent ablation study in \Cref{fig_extend}, simply populating the bank with domain-specific unlabeled data effectively rectifies these outliers.

\subsubsection{\textbf{Qualitative Comparison}}
We present a qualitative comparison of CorrCLIPv2 against training-free competitors in \Cref{fig_seg_map} across five standard benchmarks.
ClearCLIP and ProxyCLIP tend to produce spurious artifacts and noisy predictions. In contrast, CorrCLIPv2 generates semantically consistent segmentation maps. The superiority of CorrCLIPv2 is particularly evident in complex scenes. In the A150 and City, Trident and SC-CLIP exhibit noticeable misclassifications (e.g., confusing the road with the sidewalk or misidentifying indoor furniture). CorrCLIPv2, however, accurately identifies these concepts. This is attributed to the cross-image object retrieval mechanism, which effectively mitigates contextual bias. Consequently, CorrCLIPv2 maintains high semantic fidelity in complex scenarios, aligning closely with the ground truth.

\begin{table}
  \centering
  \caption{Effectiveness of each component}
  \label{tab:component}
  \begin{tabular}{l|lllll}
    \toprule
    Method & V21 & PC60 & Object & A150 & City\\
    \midrule
     Baseline & 48.1 & 29.4 & 29.4 & 18.1 & 32.7\\
     +SR & 66.7 & 40.2 & 42.0 & 24.8 & 43.8\\
     +VR & 67.1 & 40.4 & 42.5 & 24.9 & 44.2\\  
     +FR & 68.9& 42.4 & 42.5 & 26.5 & 45.2\\ 
     +MC & \textbf{75.8} & \textbf{45.9} & \textbf{45.4} & \textbf{28.6} & \textbf{53.7}\\
    \bottomrule
  \end{tabular}
\end{table}

\begin{table}
  \centering
    \caption{Impact of mask resolution on Scope Reconstruction. The ``Scale Factor'' denotes the spatial scaling ratio relative to the size of CLIP feature map}
  \label{tab:mask_size}
  \begin{tabular}{@{}c | ccccc@{}}
    \toprule
    Scale Factor & V21 & PC60 & Object& A150 & City \\
    \midrule
    No SR & 48.1 & 29.4 & 29.4 & 18.1 & 32.7 \\
    1 & 62.6 & 38.1 & 39.0 & 23.3 & 40.8 \\
    \rowcolor{mygray}
    2 & 66.7 & 40.2 & 42.0 & 24.8 & 43.8 \\
    3 & 68.0 & 40.8 & 42.7 & 25.1 & 44.8 \\
    4 & \textbf{69.2} & \textbf{41.4} & \textbf{43.5} & \textbf{25.6} & \textbf{45.6}  \\

    \bottomrule
  \end{tabular}
\end{table}

\subsection{Ablation Study on CorrCLIP}
We conduct ablation studies to validate the efficacy of each component of CorrCLIP.

\subsubsection{\textbf{Effectiveness of Each Component}}
We investigate the contribution of each proposed module: Scope Reconstruction (SR), Value Reconstruction (VR), Feature Refinement (FR), and Map Correction (MC). We employ ClearCLIP \citep{lan2024clearclip} as the baseline, which removes the feed-forward networks and residual connections in the final ViT layer. The results are summarized in \Cref{tab:component}.
Introducing SR yields the most significant performance boost. This dominant gain empirically confirms that confining attention within objects to reduce inter-class correlations improves segmentation performance.
Adding VR provides marginal yet consistent improvements across all datasets. This indicates that refining the similarity measures within the region masks helps further suppress irrelevant correlations.
FR further improves the feature representation by enhancing spatial details and region-specific semantics, while MC ensures spatial consistency. The integration of all components leads to the best performance, demonstrating their complementary nature.

\begin{table}
  \centering
    \caption{Effectiveness of the spatial branch and semantic branch}
  \label{tab:fr}
  \begin{tabular}{@{}cc|ccccc@{}}
    \toprule
    Sem. & Spa. & V21 & PC60 & Object & A150 &City\\
    \midrule
    && 67.1 & 40.4 & 42.5 & 24.9 & 44.2 \\
    \checkmark&& 67.9 & 41.2 & \textbf{42.6} & 25.8 & 43.5 \\
    &\checkmark& 68.1 & 41.9 & 41.9 & 25.9 & \textbf{45.8}\\
    \checkmark&\checkmark& \textbf{68.9} & \textbf{42.4} & 42.5 & \textbf{26.5} & 45.2 \\

    \bottomrule
  \end{tabular}
\end{table}

\begin{table*}
\small
\caption{Effectiveness of VFM-Indexed Semantic Retrieval (VISR) and Decoupled Multi-CLIP Ensemble (DMCE)\label{tab:set}}
\centering
\begin{tabular}{l|cccccccccc|c}
    \toprule
    Method & V21 & V20 & PC60 & PC59 & Object & Stuff & A150 & City & A847 & PC459 & Mean \\
    \midrule
    CorrCLIP-B & 75.8 & 89.9 & 45.9 & 50.8 & 45.4 & 32.6 & 28.6 & 53.7 & 10.5 & 11.8 & 44.2 \\
    +VISR & 78.2 & 92.2 & 49.3 & 55.5 & 48.9 & 36.9 & 32.1 & 57.4 & 10.9 & 16.0 & 47.7  \\
    +DMCE & \textbf{82.2} & \textbf{93.6} & \textbf{50.8} & \textbf{56.3} & \textbf{54.6} & \textbf{38.5} & \textbf{33.2} & \textbf{58.7} & \textbf{11.7} & \textbf{16.8} & \textbf{49.6} \\
    \midrule
    CorrCLIP-L & 79.5 & 92.2 & 47.0 & 53.1 & 52.2 & 35.4 & 32.9 & 56.0 & 14.5 & 15.2 & 47.8  \\
    +VISR & 81.3 &93.0 &49.8 &56.2 &56.1 &37.4 &35.4 &58.1 &14.8 &17.9 &50.0  \\
    +DMCE & \textbf{82.3} &\textbf{93.2} &\textbf{51.1}&\textbf{56.8} &\textbf{58.0} &\textbf{39.1} &\textbf{36.5} &\textbf{59.2} &\textbf{14.9} &\textbf{18.6} &\textbf{51.0}  \\
    \bottomrule
  \end{tabular}
\end{table*}

\begin{table*}
\small
\caption{Comparison of different VFMs used in CorrCLIPv2. DINOv2 and its distillates demonstrate superior performance due to their robust and generalizable representations \label{tab:vfm}}
\centering
\begin{tabular}{l|cccccccccc|c}
    \toprule
    VFM & V21 & V20 & PC60 & PC59 & Object & Stuff & A150 & City & A847 & PC459 & Mean \\
    \midrule
    SAM & 70.8 & 85.8 & 41.4 & 47.3 & 30.1 & 27.3 & 26.3 & 50.7 & 6.1 & 11.6 & 39.7 \\
    \midrule
    DINO& 78.0 & 90.1 & 46.9 & 51.9 & 43.4 & 32.7 & 26.7 & 51.4 & 7.2 & 13.8 & 44.2  \\
    DINOv2  & \textbf{82.7} & \textbf{93.8} & \textbf{51.0} & \textbf{56.5} & 54.3 & 38.3 & 32.5 & 57.7 & 11.1 & 15.9 & 49.4 \\
    DINOv3  & 82.4 & 93.3 & 50.4 & 56.3 & 54.0 & 38.7 & 33.1 & 58.1 & 11.7 & 16.5 & 49.5 \\
    \midrule
    PE-Spatial& 80.9 & 91.0 & 48.9 & 53.9 & 50.9 & 36.4 & 30.3 & 49.3 & 9.3 & 14.8 & 46.6 \\
    RADIOv2.5& 82.0 & 93.4 & 50.6 & 56.0 & \textbf{54.9} & \textbf{38.7} & \textbf{33.3} & \textbf{59.3} & \textbf{12.0} & 16.0 & 49.6 \\
    C-RADIOv3& 82.2 & 93.6 & 50.8 & 56.3 & 54.6 & 38.5 & 33.2 & 58.7 & 11.7& \textbf{16.8} & \textbf{49.6} \\
    \bottomrule
  \end{tabular}
\end{table*}

\subsubsection{\textbf{Impact of Mask Size in Scope Reconstruction}}

A technical challenge in SR is the resolution mismatch between the high-resolution masks and the low-resolution CLIP feature maps. Downsampling masks inevitably introduces spatial quantization errors. We hypothesize that performing SR at higher resolutions mitigates this issue. To validate this, we upsample the feature map via interpolation to match larger masks. \Cref{tab:mask_size} shows a monotonic performance increase as the scale factor grows, demonstrating that reducing quantization artifacts in SR can further improve segmentation performance. Considering the trade-off between accuracy and computational overhead, we adopt a scale factor of 2 as the default setting.

\begin{table*} 
\small
    \caption{Comparison of different datasets used to construct memory bank. 
    We report the number of images (Imgs) and embeddings (Embs) in the memory bank}
    \label{tab:dataset}
    \centering
    \begin{tabular}{lcc|cccccccccc}
        \toprule
        \multirow{2}{*}{Dataset} & \multicolumn{2}{c}{Statistics} & \multicolumn{7}{c}{Performance} & \\
        \cmidrule(lr){2-3} \cmidrule(lr){4-11}

        & \# Imgs & \# Embs & V21 & PC60 & Object & Stuff & A150 & City & A847 & PC459\\
        \midrule

         Baseline & - & - &75.8 & 45.9 & 45.4 & 32.6 & 28.6 & 53.7 & 10.5 & 11.8 \\
        \midrule
        Pascal & 17k  & 0.3M & \textbf{83.1} & \textbf{50.9} & 47.2 & 35.5 & 33.0 & \textbf{60.4} & 10.9 & 16.7  \\
        ADE20k     & 20k  & 0.6M & 82.3  & 50.6 & 45.3 & 34.9 & \textbf{33.3} & 59.7 & \textbf{12.5} & 16.1  \\  
        COCO   & 118k & 2.9M & 82.2 &50.8 & \textbf{54.6} & \textbf{38.5} & 33.2 & 58.7 & 11.7 & \textbf{16.8}  \\
        \bottomrule
    \end{tabular}
\end{table*}

\begin{table}
  \centering
  \caption{Impact of different mask generators}
  \label{tab:mask_generator}
  \begin{tabular}{@{}l|ccccc@{}}
    \toprule
    Mask Generator & V21 & PC60 & Object & A150 & City \\
    \midrule
    EoMT & 81.9 & \textbf{51.0} & \textbf{56.5} & 29.2 & 56.5  \\
    SAM & 79.7 & 47.6 & 51.2 & 30.2 & 53.5\\
    SAM2 & 80.8 & 48.7 & 53.2 & 30.0 & 56.7  \\
    EntitySeg & \textbf{82.2} & 50.8 & 54.6 & \textbf{33.2} & \textbf{58.7} \\
    \midrule
    Ground Truth & 91.9 & 59.5 & 76.1 & 46.1 & 77.6  \\
    \bottomrule
  \end{tabular}
\end{table}

\subsubsection{\textbf{Ablation for Feature Refinement}}
We examine the individual and combined effects of the Spatial Branch and Semantic Branch within the Feature Refinement module in \Cref{tab:fr}.
While both branches aim to enhance feature representation, applying them in isolation yields mixed results. For instance, using only the Semantic Branch or Spatial Branch leads to slight performance drops on City or Object. This is attributed to potential misalignment, as modifying features with a single type of prior may disrupt the delicate alignment between visual and text embeddings. When combined, the two branches function synergistically, resulting in improvements across all benchmarks.

\subsection{Ablation Study on CorrCLIPv2}
Here, we conduct comprehensive ablation studies to present the effectiveness of CorrCLIPv2.

\subsubsection{\textbf{Effectiveness of Semantic Retrieval and Ensemble Strategies}}
We evaluate the contributions of the proposed cross-image extension modules: VFM-Indexed Semantic Retrieval and Decoupled Multi-CLIP Ensemble. The breakdown of results is shown in \Cref{tab:set}.
Integrating the retrieval mechanism yields substantial and consistent improvements across all ten benchmarks. For the CorrCLIP-B variant, this module boosts the averaged mIoU from 44.2\% to 47.7\%. 
This empirical evidence strongly supports our premise that aggregating semantically consistent instances from a diverse memory bank can distill a semantic prototype that rectifies the semantic drift. The trend holds for the larger CorrCLIP-L model, confirming the scalability of our retrieval paradigm.

Additionally incorporating the Decoupled Multi-CLIP Ensemble further improves performance. For CorrCLIP-B, the averaged mIoU increases to 49.6\%, and for CorrCLIP-L, it reaches a state-of-the-art 51.0\%. This indicates that fusing representations from CLIP models pre-trained on different distributions enriches the semantic space, making the prediction more stable. Crucially, unlike traditional ensembles that double the computational cost, our decoupled strategy integrates these priors into the memory bank offline, achieving these gains with negligible marginal cost during inference.

\subsubsection{\textbf{Impact of Different Vision Foundation Models}}
The core of our retrieval mechanism lies in the quality of the feature space used to index the memory bank. We investigate the impact of different VFMs. We categorize representative VFMs into three types: (1) Segmentation Models (SAM \citep{sam}); (2) SSL Models (DINO series \citep{dino, dinov2, dinov3}); and (3) Agglomerative Models that fuse multiple priors (PE-Spatial \citep{pe} and RADIO series \citep{radiov2}). The comparative results are detailed in \Cref{tab:vfm}.
SAM performs poorly as its latent space prioritizes geometric primitives over semantic concepts, leading to inconsistent retrieval. PE-Spatial offers partial improvements by integrating SAM and CLIP but still falls short of SSL models. In the SSL models, DINOv2 significantly outperforms the ImageNet-trained DINO, highlighting the importance of massive-scale pre-training for robust feature extraction. RADIO achieves the best results by distilling complementary priors from multiple models. The results demonstrate that SSL-based VFMs are most suitable for CorrCLIPv2.

\subsubsection{\textbf{Impact of Different Datasets for Memory Bank}}
We investigate the impact of the source dataset used to construct the memory bank on the final segmentation performance. We evaluate the performance of three distinct datasets: Pascal VOC, ADE20K, and COCO. 
As shown in \Cref{tab:dataset}, even smaller-scale datasets yield significant performance gains over the single-image baseline. This validates that the mechanism of cross-image retrieval is fundamentally robust, functioning effectively even with limited support data. 

\begin{figure*}[t]
\centering
\includegraphics[width=1\linewidth]{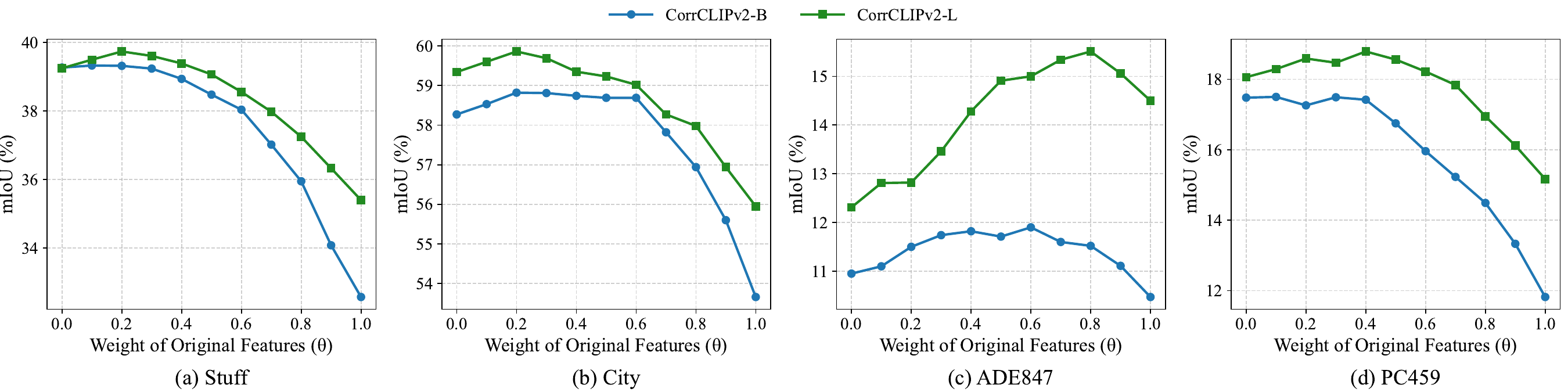}
\caption{Impact of the fusion weight $\theta$. The original and the retrieved support features are aggregated with weights $\theta$ and $1-\theta$, respectively. The optimal performance is observed in the range of $[0.2, 0.8]$. We set the default value to 0.5}
\label{fig_weight}
\end{figure*}

\begin{figure}[t]
\centering
\includegraphics[width=0.8\linewidth]{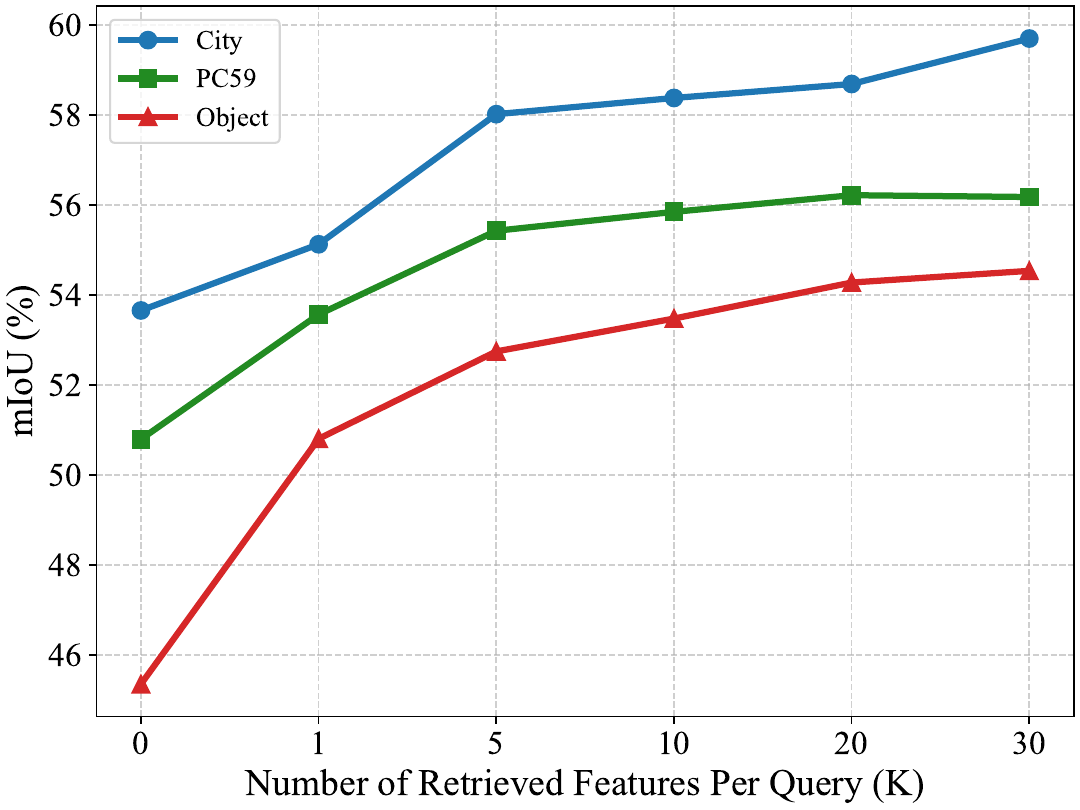}
\caption{Impact of the number of retrieved instances for each query ($K$). Performance generally improves with more support instances and saturates around $K=20$}
\label{fig_num_refer}
\end{figure}

\subsubsection{\textbf{Impact of Different Mask Generators}}
The quality of mask proposals is foundational to our framework. We evaluate different class-agnostic mask generators in \Cref{tab:mask_generator}.
EntitySeg \citep{entityseg} yields superior performance, significantly outperforming prompt-based SAM. This advantage stems from the fact that EntitySeg is explicitly optimized for open-world entity segmentation, producing distinct, non-overlapping masks that perfectly satisfy the requirements of our framework. In contrast, SAM, which relies on grid-point prompting, inevitably generates redundant and overlapping masks, introducing ambiguity. EoMT \citep{eomt} exhibits limited generalization capabilities. Substituting generated proposals with ground-truth masks yields substantial performance gains. These results suggest that CorrCLIPv2 effectively capitalizes on high-quality proposals, and its performance will naturally scale with future advancements in class-agnostic segmentation.

\begin{figure*}[t]
\centering
\includegraphics[width=1\linewidth]{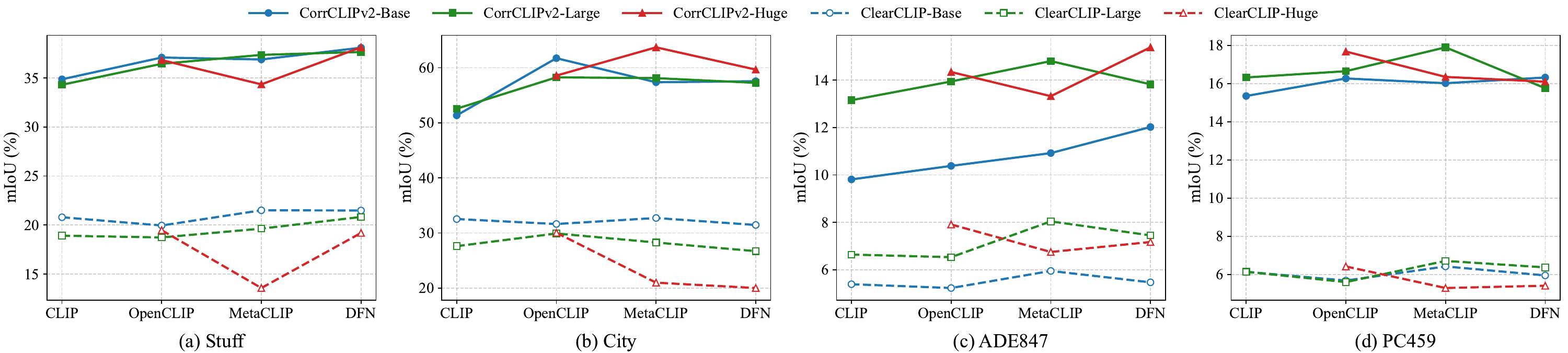}
\caption{Performance of CorrCLIPv2 across diverse CLIP backbones. We evaluate performance using various model capacities and pre-training sources. The consistent improvement over baselines across all configurations demonstrates that our framework is backbone-agnostic, effectively enhancing arbitrary CLIP models regardless of their specific pre-training priors}
\label{fig_clip}
\end{figure*}

\subsubsection{\textbf{Ablation Study on Retrieval Hyperparameters}}
We analyze two critical hyperparameters governing the retrieval mechanism: the number of retrieved instances for each query ($K$) and the fusion weight ($\theta$), which controls the trade-off between the original object feature and the retrieved set. 
As shown in \Cref{fig_num_refer}, segmentation performance improves monotonically as $K$ increases, benefiting from a more robust semantic prototype. The gain saturates around $K=20$, which we adopt to balance accuracy and efficiency.
The impact of $\theta$ is shown in \Cref{fig_weight}. The optimal performance is observed within the range of $[0.2, 0.8]$. We select $\theta=0.5$ as the default to mitigate contextual bias via retrieval while preserving necessary original object contexts.

\begin{table}
  \caption{Computational costs on RTX 4090 with FP16. Performance is mean mIoU across ten standard benchmarks. CorrCLIPv2-fast uses smaller mask generator and VFM}
  \label{tab:cost}
  \centering
  \begin{tabular}{@{}l|ccc|c@{}}
    \toprule
    \multirow{2}{*}{Method} & Latency & Memory & Para. & Perf. \\ 
                            & (ms)    & (GB)   & (M)       & (mIoU)      \\ 
    \midrule
    {ClearCLIP } & 19 & 0.7 & 150 & 31.9 \\
    {ProxyCLIP } & 69 & 1.9 & 235 & 36.7\\
    {Trident } & 81 & 2.5 & 364 & 41.0 \\
    \midrule
    CorrCLIP & 177 & 3.0 & 458 & 43.1 \\
    CorrCLIPv2& 151 & 4.3 & 756 & 49.6 \\
    CorrCLIPv2-fast & 54 & 2.0 & 335 & 46.6 \\
    \bottomrule
  \end{tabular}
\end{table}

\subsubsection{\textbf{Computational Costs}}
\Cref{tab:cost} presents a comprehensive comparison of computational costs.
Despite its expanded architecture, CorrCLIPv2 achieves lower latency than the original CorrCLIP, attributed to the efficient EntitySeg mask generator and the negligible overhead of the retrieval extension. To address the increased memory footprint, we introduce CorrCLIPv2-fast, a variant utilizing lightweight alternatives (EoMT-S and C-RADIO-B). This version drastically reduces resource consumption while retaining impressive performance. Notably, it achieves an averaged mIoU of 46.6\%, significantly outperforming Trident and ProxyCLIP while running faster. This suggests that our framework's superiority stems from its robust design logic rather than sheer model scaling.

\subsubsection{\textbf{Ablation Study on CLIP}}
We assess the generalizability of CorrCLIPv2 across various CLIP variants, differing in model capacity and pre-training data sources, without employing the ensemble strategy. As illustrated in \Cref{fig_clip}, our framework consistently yields substantial performance gains over the baselines across all configurations. This confirms that our contributions are backbone-agnostic and scalable, effectively enhancing the discriminability of arbitrary CLIP models regardless of their specific pre-training priors.

\begin{figure}[t]
\centering
\includegraphics[width=1\linewidth]{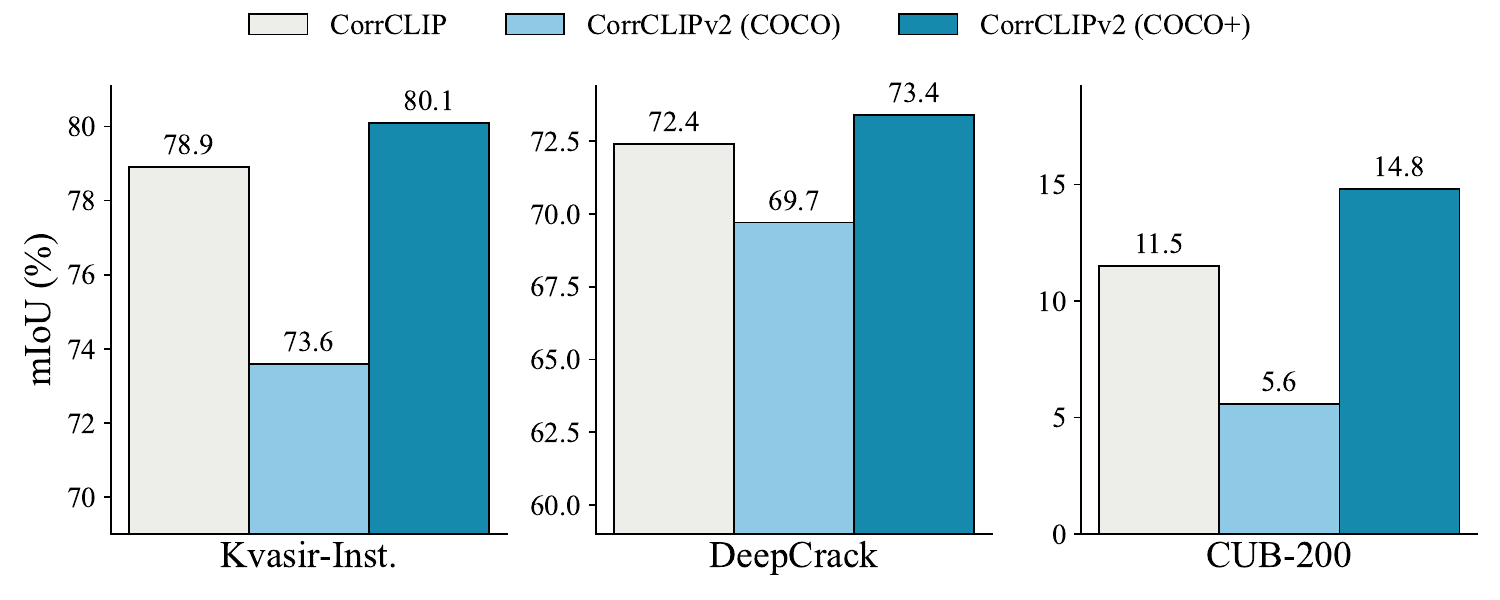}
\caption{Demonstration of the memory bank's dynamic extensibility. By simply augmenting the COCO index with domain-specific unlabeled images (COCO+), CorrCLIPv2 achieves immediate performance gains on specialized benchmarks. This validates our framework's efficient, plug-and-play adaptation capability}
\label{fig_extend}
\end{figure}

\subsubsection{\textbf{Extensibility of the Memory Bank}}
A distinct advantage of our retrieval-based paradigm is its flexible extensibility. Unlike training-based models that require retraining to absorb new data, our memory bank can be updated on-the-fly. To demonstrate this, we augment the base COCO memory bank with unlabeled images from the three specialized datasets where performance drops: {Kvasir-Instrument}, {DeepCrack}, and {CUB-200}. As illustrated in \Cref{fig_extend}, incorporating domain-specific samples yields immediate and significant performance gains. This result validates that CorrCLIPv2 can adapt to highly specialized domains simply by adding relevant raw data, offering a flexible solution for continuous adaptation.

\begin{table}
  \centering
  \caption{Comparison of different semantic features for constructing the memory bank. Using CorrCLIP to extract semantic features yields the best results, validating the effectiveness of correlation reconstruction mechanism}
  \label{tab:ovss}
  \begin{tabular}{@{}l|ccccc@{}}
    \toprule
    Method & V21 & PC60 & Object & A150 & City\\
    \midrule
    CLIP  & 11.6 & 5.0 & 1.7 & 1.9 & 3.0 \\
    ClearCLIP & 77.5 & 45.4 & 53.0 & 28.3 & 41.0 \\
    ProxyCLIP & 78.5 & 46.8 & 54.5 & 30.5 & 51.1 \\
    CorrCLIP & \textbf{82.1} & \textbf{50.8} & \textbf{56.9} & \textbf{33.4} & \textbf{58.3} \\
    \bottomrule
  \end{tabular}
\end{table}

\subsubsection{\textbf{Ablation Study on Semantic Features in Memory Bank}}
We investigate how the quality of the semantic features stored in the memory bank affects the final performance. We vary the model used to extract the semantic features: standard CLIP \citep{clip}, ClearCLIP \citep{lan2024clearclip}, ProxyCLIP \citep{proxyclip}, and our CorrCLIP. The results in \Cref{tab:ovss} reveal a clear hierarchy. Standard CLIP yields the lowest performance, as its incoherent patch features fail to provide accurate semantics. ClearCLIP and ProxyCLIP achieve respectable gains, validating that our retrieval framework is fundamentally effective given reasonable support features. Crucially, CorrCLIP delivers the best performance. This demonstrates the superiority of CorrCLIP for improving CLIP's features for segmentation.

\begin{table}
  \centering
  \caption{Performance of CorrCLIPv2 under oracle text prompts}
  \label{tab:oracle}
  \begin{tabular}{@{}l|cccccc@{}}
    \toprule
    Method & PC59 & Stuff & A150 & A847 & PC459\\
    \midrule
    CorrCLIPv2 & 56.8 & 39.1 & 36.5 & 14.9 &  18.6\\
    Oracle Prompts& 74.0 & 54.3 & 58.1 & 35.8 & 52.2\\
    \bottomrule
  \end{tabular}
\end{table}

\subsection{Limitations and Future Work}
In open-vocabulary settings, the presence of numerous irrelevant categories inevitably introduces noise and semantic ambiguity. We conduct an oracle prompt experiment where we restrict text queries exclusively to categories actually present in the image. As shown in \Cref{tab:oracle}, the significant performance gap between this ideal scenario and standard inference highlights the model’s susceptibility to semantic distractors from a massive open-ended vocabulary.
Therefore, a critical avenue for future work is the development of robust image-level category pre-filtering strategies. Specifically, prior to pixel-level segmentation, we can employ auxiliary methods to determine which categories are likely present in the scene. This could be achieved by leveraging lightweight multi-label classification networks, calculating global image-text similarity scores, or prompting large vision-language models to generate a set of candidate objects. By implementing such a recognize-then-segment paradigm, we can effectively prune irrelevant text queries, mitigate semantic distractors, and further narrow the gap toward oracle performance.

\section{Conclusion}\label{sec13}
In this work, we reveal that inter-class patch correlations in CLIP primarily manifest as noise, hindering segmentation performance. To address this, we propose a unified, training-free object-centric framework, named CorrCLIPv2. By explicitly disentangling inter-class interactions, strengthening feature representations, and unifying intra-mask predictions, our approach significantly enhances CLIP's segmentation capabilities.
Furthermore, to mitigate contextual bias, we evolve our framework into a cross-image paradigm. We introduce VFM-indexed semantic retrieval to accurately access semantically consistent support features from a large-scale memory bank of unlabeled data. This retrieval mechanism, combined with our decoupled multi-CLIP ensemble strategy, significantly enhances prediction accuracy with minimal inference overhead. Extensive experiments demonstrate that CorrCLIPv2 sets a new state-of-the-art across ten standard benchmarks and notably outperforms fully supervised open-vocabulary counterparts on the MESS benchmark. We believe this work provides a scalable, robust foundation for open-vocabulary semantic segmentation.

\backmatter

\bibliography{sn-bibliography}%

\end{document}